\newif\ifshowbio
\showbiofalse

\newif\ifconfver
\confverfalse      

\newif\ifplainver  
\plainvertrue

\ifplainver
    \confverfalse   
\fi

\ifconfver
     \documentclass[10pt,twocolumn,twoside]{IEEEtran}
\else
    \ifplainver
        \documentclass[11pt]{article}
        \usepackage{fullpage}
    \else
        \documentclass[12pt,draftcls,onecolumn]{IEEEtran}
    \fi
\fi

\usepackage{calc,amsfonts,amssymb,amsmath,bm,url,color,theorem,graphicx,cite}
\usepackage{psfrag,subfigure,float}
\usepackage[algoruled,linesnumbered]{algorithm2e}
\usepackage{multirow}
\usepackage{soul}

\definecolor{orange}{RGB}{255,107,0}

\newtheorem{Fact}{Fact}
\newtheorem{Lemma}{Lemma}

\newtheorem{Theorem}{Theorem}
\newtheorem{Def}{Definition}
\newtheorem{Corollary}{Corollary}

\newtheorem{Observation}{Observation}
\theorembodyfont{\rmfamily}

\newtheorem{Assumption}{Assumption}
\newtheorem{Remark}{Remark}

\begin{document}

\bibliographystyle{IEEEtran}

\newcommand{\papertitle}{
Self-Dictionary Sparse Regression for Hyperspectral Unmixing: Greedy Pursuit and Pure Pixel Search are Related
}

\newcommand{\paperabstract}{
This paper considers a recently emerged hyperspectral unmixing formulation based on sparse regression of a self-dictionary multiple measurement vector (SD-MMV) model,
wherein the measured hyperspectral pixels are used as the dictionary.
Operating under the pure pixel assumption,
this SD-MMV formalism is special in
that it allows
simultaneous identification of the endmember spectral signatures and the number of endmembers.
Previous SD-MMV studies mainly focus on convex relaxations.
In this study, we explore the alternative of greedy pursuit, which generally provides efficient and simple algorithms.
In particular, we design a greedy SD-MMV algorithm using simultaneous orthogonal matching pursuit.
Intriguingly, the proposed greedy algorithm is shown to be closely related to some existing pure pixel search algorithms, especially, the successive projection algorithm (SPA).
Thus, a link between SD-MMV and pure pixel search is revealed.
We then perform exact recovery analyses,
and prove that the proposed greedy algorithm is robust to noise---including its identification of the (unknown) number of endmembers---under a sufficiently low noise level.
The identification performance of the proposed greedy algorithm is demonstrated through both
synthetic and real-data experiments.
}


\ifplainver

    \date{September, 2014, Revised, March, 2015}

    \title{\papertitle\thanks{Part of this work was published in EUSIPCO 2013 \cite{fu2013greedy}.}}

    \author{
    $^\dag$Xiao Fu, $^\dag$Wing-Kin Ma, $^\ddag$Tsung-Han Chan, and $^*$Jos\'{e} M. Bioucas-Dias
    \\ ~ \\
    $^\dag$Department of Electronic Engineering, The Chinese University of Hong  Kong, \\
    Hong Kong \\
    Email: wkma@ieee.org, xfu@ee.cuhk.edu.hk
    \\ ~ \\
    $^\ddag$ Sunplus Technology Co., Taiwan \\
    Email: thchan@ieee.org
    \\ ~ \\
    $^*$ Instituto de Telecomunica\c{c}\~{o}es and
    Instituto Superior T\'{e}cnico, \\
    University of Lisbon, 1049-1, Lisbon, Portugal
     \\
    Email: bioucas@lx.it.pt
    }

    \maketitle

    \begin{abstract}
    \paperabstract
    \end{abstract}

\else
    \title{\papertitle}

    \ifconfver \else {\linespread{1.1} \rm \fi

   \author{Xiao Fu, Wing-Kin Ma, Tsung-Han Chan, and Jos\'{e} M. Bioucas-Dias
\thanks{
Copyright (c) 2014 IEEE. Personal use of this material is permitted. However, permission to use this material for any other purposes must be obtained from the IEEE by sending a request to pubs-permissions@ieee.org.

This work was supported
in part by General Research Funds of Hong Kong Research Grant Council (Project Nos. CUHK415509 and CUHK4441269),
and in part
by the Portuguese Science and Technology
Foundation under Project UID/EEA/50008/2013, Project PTDC/EEIPRO/
1470/2012. Part of this work was published in EUSIPCO 2013 \cite{fu2013greedy}.

X. Fu was with the Department of Electronic Engineering, the Chinese University of Hong Kong, Shatin, N.T., Hong Kong;
he is now with the Department of Electrical and Computer Engineering, University of Minnesota, Minneapolis, MN 55455 USA (e-mail: xfu@umn.edu).
W.-K. Ma is with the Department of Electronic Engineering, the Chinese University of Hong Kong, Shatin, N.T., Hong Kong (e-mail: wkma@ee.cuhk.edu.hk).
T.-H. Chan is with the MediaTek Inc., Hsinchu, Taiwan (e-mail: thchan@ieee.org).
J.M. Bioucas-Dias is with the Instituto de Telecomunica\c{c}\~{o}es and Instituto Superior T\'{e}cnico,
    University of Lisbon, 1049-1, Lisbon, Portugal (e-mail: bioucas@lx.it.pt).}
}

    \maketitle

    \ifconfver \else
        \begin{center} \vspace*{-2\baselineskip}
        \end{center}
    \fi

    \begin{abstract}
    \paperabstract
    \end{abstract}

    \begin{keywords}\vspace{-0.0cm}
        Greedy pursuit, self-dictionary sparse regression, hyperspectral unmixing, number of endmembers estimation
    \end{keywords}

    \ifconfver \else \IEEEpeerreviewmaketitle} \fi

 \fi

\ifconfver \else
    \ifplainver \else
        \newpage
\fi \fi

\section{Introduction}

Hyperspectral unmixing (HU) aims at determining the spectra of the underlying endmembers (or materials) and the corresponding proportions in each sensed pixel from a captured hyperspectral image (HSI).
It is an important branch of techniques in hyperspectral data analysis and processing,
enables many applications in remote sensing~\cite{bioucas2013hyperspectral},
and has close relationship to topics in other contexts, such as non-negative matrix factorization (NMF) in machine learning~\cite{gillis2014why}.
Readers are referred to the literature, such as \cite{Bioucas2012,Ma2014} and the references therein, for descriptions of various HU approaches.

Recently, a new sparse optimization-based approach was proposed for HU \cite{Esser2012};
see also \cite{Elhamifar2012,gillis2014why,recht2012factoring,gillis2013robustness,gillis2014robust} for other contexts.
This approach uses the measured pixel vectors themselves as the (overcomplete) dictionary to perform sparse basis selection.
By doing so, a smallest subset of the measured pixel vectors for representing all the measured pixel vectors is sought,
and the obtained pixel subset
is taken as the endmember spectra estimates.
Such a self-dictionary multiple measurement vector (SD-MMV) formulation is unlike those seen in the currently active developments in dictionary-aided sparse regression~\cite{Iordache2011,Iordache2012collaborative,joseMUSIC2013}, wherein a spectral library is often provided in advance, and endmember estimates are selected from the library.
Instead, the SD-MMV approach is closer to the pure pixel search approach~\cite{PPI,NFINDR,VCA,MC01,ATGP,VMAX}
 in terms of rationale.
In essence, it is shown that finding the aforementioned pixel subset in SD-MMV amounts to identifying pure pixels~\cite{Esser2012}; i.e., pixels that contain only one endmember.
SD-MMV is more than just an alternative means for pure pixel search, however.
Its formulation encourages one to perform simultaneous identification of the pure pixels and the number of endmembers---which is attractive since we generally require a separate model order estimator~\cite{Hysime,HFC} to identify the number of endmembers, and such a problem is generally challenging to cope with.
In comparison,
the problem of simultaneous identification of the pure pixels and the number of endmembers
is less considered in the traditional pure pixel search studies;
see \cite{GENE,ATGP-NPD} for recent works that start to look at this direction.

At present,
most of the studies handle the sparse SD-MMV optimization problem via
convex relaxations~\cite{Esser2012,Elhamifar2012,recht2012factoring,gillis2013robustness,gillis2014robust}.
Remarkably, it is shown that convex SD-MMV relaxations can provide exact recovery of the pure pixels if pure pixels exist and noise is absent~\cite{Esser2012,Elhamifar2012}; see \cite{recht2012factoring,gillis2013robustness,gillis2014robust} for further results in the noisy case.
However, convex SD-MMV relaxations have a drawback---its computational overheads are generally high, since the number of optimization variables involved in a convex SD-MMV relaxation is the square of the number of pixels
(e.g., an HSI with
$1,000$
pixels amounts to
$1,000,000$
optimization variables).
The existing works circumvent this issue by downsizing the HSI data through extra processing~\cite{Esser2012}; see also~\cite{gillis2014why}.

In this paper, we tackle the SD-MMV HU problem using a different strategy, namely, greedy pursuit.
Greedy pursuit is a well-known and frequently-employed tool for handling compressive sensing
or sparse optimization problems \cite{Tropp2006}, besides convex relaxations.
It generally leads to
algorithms that are computationally much cheaper than those by convex relaxations, and
thus is believed to be more suitable for `big data' problems such as the HU problem.
Our study has its emphasis on easy-to-implement algorithms, while establishing theoretical grounds for their soundness at the same time.
Our contributions are summarized below.

\begin{enumerate}
\item We consider the application of a popular greedy method, namely, simultaneous orthogonal matching pursuit (SOMP), to the SD-MMV formulations.
    The resulting algorithm, called SD-SOMP for short, is shown to be closely related to some existing pure pixels search algorithms, namely, successive projection algorithm (SPA) \cite{MC01,Gillis2012}, automatic target generation process (ATGP) \cite{ATGP}, and successive volume maximization (SVMAX) \cite{VMAX}.
    Such a connection is not seen in convex SD-MMV relaxations,
    and is interesting in the sense of building a link between the long-existing pure pixel search algorithm class and the relatively new sparse regression developments.

\item We proceed further by studying the noisy case. Armed with a recent theoretical result by Gillis and Vavasis \cite{Gillis2012}, we analyze exact recovery conditions of SD-SOMP in the noisy case and under unknown number of endmembers. The analysis shows that SD-SOMP is robust against noise perturbations under sufficiently high signal-to-noise ratios (SNRs)---including the identification of the number of endmembers. We also prove exact recovery conditions of the original (and generally much harder) SD-MMV  problem under the same settings, which is not only meaningful for the purpose of understanding the relative approximation accuracy of SD-SOMP in the present work, but could also be of independent interest in other contexts such as NMF.
\end{enumerate}

In the conference version of this paper \cite{fu2013greedy},
the first contribution was presented.
The second contribution is new; in fact, it represents a more significant contribution of this paper.

We should mention that in one of our SD-SOMP designs, the algorithm is observed to bear some resemblance to a few existing pure pixel search
works \cite{GENE,ATGP-NPD} that consider simultaneous identification of pure pixels and the the number of endmembers; the
similarity lies in the stopping criterion.
What sets this work apart is that our design is equipped with rigorous
exact recovery analysis.
In fact, numerical and real-data experiments will show that under high
SNRs, our proposed design provides consistent and generally better estimation performance than
the previous works.

\section{Background}

\subsection{Notation}

We use notations commonly seen in signal processing, HU or related fields.
The notations ${\bf x} \in \mathbb{R}^n$ and ${\bf X} \in \mathbb{R}^{m \times n}$ mean that ${\bf x}$ and ${\bf X}$ are a real-valued $n$-dimensional vector and a real-valued $m \times n$ matrix, respectively (resp.).
The superscript ``$T$'' stands for transpose.
The vector $\ell_q$ norm, $q \geq 1$, is denoted by $\| \cdot \|_q$.
In addition,
the $i$th column of a matrix ${\bf X} \in \mathbb{R}^{m \times n}$ is denoted by ${\bf x}_i \in \mathbb{R}^m$;
the notation ${\bf x} \geq {\bf 0}$ (resp. ${\bf X} \geq {\bf 0}$) means that ${\bf x}$ (resp. ${\bf X}$) is element-wise non-negative;
${\bf 1}$ denotes an all-one vector;
${\bf e}_k$ denotes a unit vector where $[ {\bf e}_k ]_k = 1$ and $[ {\bf e}_k ]_i = 0$ for all $k \neq i$;
the cardinality of a given discrete set $\Lambda$ is denoted by $|\Lambda|$;
given a matrix ${\bf X} \in \mathbb{R}^{m \times n}$ and an index set $\Lambda \subset \{ 1, \ldots, n \}$,
${\bf X}_{\Lambda} \in \mathbb{R}^{m \times |\Lambda|}$ denotes a submatrix of ${\bf X}$ where we choose a subset of columns of ${\bf X}$ whose indices are listed in $\Lambda$;
similarly ${\bf X}_{1:k} = [~ {\bf x}_1, \ldots, {\bf x}_k ~]$ denotes a submatrix of ${\bf X}$ that contains the first $k$ columns of ${\bf X}$;
${\rm rowsupp}({\bf X}) = \{ j ~|~ {\bf x}^j \neq {\bf 0} \}$ denotes the row support of ${\bf X}$, where ${\bf x}^j$ denotes the $j$th row of ${\bf X}$;
$\| {\bf X} \|_{\rm row-0} = | {\rm rowsupp}({\bf X}) |$ denotes the row-$\ell_0$ quasi-norm, or the number of nonzero rows, of ${\bf X}$;
${\bf P}^\perp_{\bf X} = {\bf I} - {\bf X} ( {\bf X}^T {\bf X})^\dag {\bf X}^T$ is the orthogonal complement projector of ${\bf X}$, where the superscript ``$\dag$'' stands for the pseudo-inverse;
and
$\sigma_{\min}({\bf X})$ and $\sigma_{\max}({\bf X})$ denote the smallest and largest singular values of ${\bf X}$, resp.

\subsection{Problem Setup}
\label{sec:problem_setup}

The problem setting in this paper is identical to that of the widely-adopted linear mixing model.
Readers are referred to the literature, such as \cite{Bioucas2012,Ma2014}, for detailed coverage,
and herein we concisely state the model.
Specifically, we have
\begin{equation} \label{eq:model}
{\bf x}[n] = {\bf A} {\bf s}[n] + {\bf v}[n], \quad n=1,\ldots,L,
\end{equation}
where
${\bf x}[n] \in \mathbb{R}^M$ denotes the measured hyperspectral pixel vector at pixel $n$, with $M$ being the number of spectral bands;
${\bf A} = [~ {\bf a}_1, \ldots, {\bf a}_N ~] \in \mathbb{R}^{M \times N}$ is the endmember signature matrix, in which
each column ${\bf a}_i$ is the hyperspectral signature vector of a distinct endmember
and $N$ is the number of endmembers;
${\bf s}[n] \in \mathbb{R}^N$ is the abundance vector at pixel $n$;
${\bf v}[n] \in \mathbb{R}^M$ is noise;
$L$ is the number of endmembers.
Every ${\bf s}[n]$ is assumed to satisfy the non-negativity and sum-to-one constraints, i.e., ${\bf s}[n] \geq {\bf 0}$ and ${\bf 1}^T {\bf s}[n] = 1$, resp., and
the columns of ${\bf A}$ are assumed to be linearly independent.
For convenience, we will let ${\bf X}= [~ {\bf x}[1], \ldots, {\bf x}[L] ~]$.


In HU, the problem is to estimate the endmember signatures ${\bf a}_1, \ldots, {\bf a}_N$ and the abundance vectors ${\bf s}[1], \ldots, {\bf s}[L]$ from the measured dataset $\{ {\bf x}[n] \}_{n=1}^L$.
We will focus only on estimation of endmember signatures;
once ${\bf A}$ is acquired, ${\bf s}[n]$'s can be recovered by standard non-blind unmixing methods; e.g., \cite{heinz2001fully}.
A common assumption in many HU studies is that the number of endmembers $N$ has been estimated {\em a priori} (specifically, by another algorithm such as \cite{Hysime,HFC}), and presumably, perfectly estimated.
This work will consider estimation of both the endmember signatures and the number of endmembers.

It is important to introduce the concepts of pure pixels~\cite{PPI,NFINDR,VCA,MC01,ATGP,VMAX,Ma2014}, since the framework to be presented is strongly connected to such concepts.
In particular, to accurately describe some theoretical results shown later,
some precise definitions for pure pixels are essential.

\begin{Def}
An index $n \in \{ 1,\ldots,L \}$ is called a pure pixel index of endmember $k$ if
${\bf s}[n] = {\bf e}_k$.
\end{Def}

\begin{Def}
The pure pixel assumption is said to hold if, for all $k \in \{1, \ldots, N \}$,
a pure pixel index of endmember $k$ exists.
\end{Def}

\begin{Def}
Suppose that the pure pixel assumption holds.
An index set $\Lambda \subset \{ 1, \ldots, L \}$ is called a complete pure pixel index set, or simply complete, if $\Lambda$ contains a pure pixel index of every endmember and $| \Lambda | = N$.
\end{Def}

Also, for ease of explanation, the number of endmembers $N$ will alternatively be called {\em model order} in the sequel.
The pure pixel assumption physically translates into scenarios where
some measured pixels are constituted purely by one endmember,
and every endmember has such pure pixels in the captured scene.
Note that the pure pixel indices are unknown,
and that one endmember may have more than one pure pixel index.
Under the pure pixel assumption, the HU problem may be formulated as that of identifying a complete pure pixel index set $\Lambda$~\cite{PPI,NFINDR,VCA,MC01,ATGP,VMAX}.
Let us briefly review this by considering the noiseless case.
By letting $n_k$ be a pure pixel index of endmember $k$ (and assuming its existence),
the measured hyperspectral vector at pixel $n_k$ equals ${\bf x}[n_k]= {\bf A}{\bf e}_k = {\bf a}_k$.
Hence,
if we can secure a collection of the indices $n_1, \ldots, n_N$,
or equivalently, a complete pure pixel index set $\Lambda= \{ n_1, \ldots, n_N \}$,
then ${\bf X}_{\Lambda}= [~ {\bf x}[n_1],\ldots, {\bf x}[n_N] ~]$ is the true endmember signature matrix ${\bf A}$ up to a column permutation.
Note that in the noisy case, ${\bf X}_{\Lambda}$ becomes a noise perturbed version of ${\bf A}$;
in practice, such an estimate may be reasonable at least for high SNRs.
Also, by saying identification of a complete $\Lambda$,
we may refer to pure pixel index identification without knowledge of $N$---that is, simultaneous estimation of the endmember signatures and the model order.


\subsection{Multiple Measurement Vector Model}
\label{sec:MMV_review}

To understand the principle of self-dictionary sparse regression for HU, it would be helpful to start with the multiple measurement vector (MMV) model in compressive sensing~\cite{Chen2006,Tropp2006,Tropp2006pt2,Eldar2010average};
a concise review is as follows.
We are given a multitude of measurement vectors ${\bf x}[n] \in \mathbb{R}^M$, $n=1,\ldots,L$.
The problem is to represent each ${\bf x}[n]$ by a linear combination of some atoms taken from a dictionary,
and one wishes to do so with the fewest number of atoms.
Let ${\bf B} \in \mathbb{R}^{M \times K}$ be the dictionary where each column ${\bf b}_i$ is an atom and $K$ is the dictionary size,
and ${\bf X}= [~ {\bf x}[1], \ldots, {\bf x}[L] ~] \in \mathbb{R}^{M \times L}$.
We aim at performing the representation
\begin{equation} \label{eq:X_BC}
{\bf X} = {\bf B}{\bf C}
\end{equation}
for some coefficient matrix ${\bf C} \in \mathbb{R}^{K \times L}$.
By letting $\Lambda = {\rm rowsupp}({\bf C})$,
which lists all indices of nonzero rows of ${\bf C}$,
it is easy to see that the active atoms in \eqref{eq:X_BC}  are $\{ {\bf b}_i \}_{i \in \Lambda}$.
Thus, selecting a minimal number of atoms for \eqref{eq:X_BC} is the same as minimizing the number of nonzero rows of ${\bf C}$ in \eqref{eq:X_BC}.
This observation leads to the following formulation for sparse MMV representation
\begin{equation} \label{eq:MMV}
\begin{aligned}
\min_{ {\bf C} \in \mathbb{R}^{K \times L}}  & ~  \| {\bf C} \|_{\rm row-0}   \\
{\rm s.t.}  &  ~  {\bf X} = {\bf B} {\bf C};
\end{aligned}
\end{equation}
see \cite{Chen2006,Tropp2006,Tropp2006pt2,Eldar2010average}.


\subsection{Self-Dictionary MMV}
\label{sec:SD-MMV}

A novel sparse MMV formulation was recently introduced in \cite{Esser2012,Elhamifar2012}.
The idea there is to use the measured dataset ${\bf X}$ itself as the dictionary.
By doing so, one seeks to find a smallest subset of measurement vectors to represent the whole set of measurement vectors.
Following the MMV formulation in the last subsection,
an SD-MMV formulation may be written as
\begin{equation} \label{eq:SD-MMV}
\begin{aligned}
\min_{ {\bf C} \in \mathbb{R}^{L \times L}}  & ~  \| {\bf C} \|_{\rm row-0}   \\
{\rm s.t.}  &  ~  {\bf X} = {\bf X} {\bf C}, ~ {\bf C} \geq {\bf 0}, ~ {\bf 1}^T {\bf C} = {\bf 1}^T;
\end{aligned}
\end{equation}
see \cite{Elhamifar2012}.
Note that the SD-MMV formulation above sets the dictionary as ${\bf B}= {\bf X}$;
cf.~\eqref{eq:X_BC}.
Moreover, the SD-MMV formulation introduces two additional model constraints,
namely, ${\bf C} \geq {\bf 0}$ and ${\bf 1}^T {\bf C} = {\bf 1}^T$.
The two constraints
mean that we model each measurement vector ${\bf x}[n]$ as a {\em convex} combination of the atoms $\{ {\bf x}[i] \}_{i=1}^L$;
specifically, by noting that ${\bf c}_n$ is the $n$th column of ${\bf C}$,
the constraints in \eqref{eq:SD-MMV} can be equivalently written as
${\bf x}[n] = {\bf X}{\bf c}_n$, ${\bf c}_n \geq {\bf 0}$, ${\bf 1}^T {\bf c}_n = 1$, for all $n$.

Our interest in the SD-MMV model lies in the HU problem.
Consider the linear hyperspectral signal model in \eqref{eq:model} in the noiseless case, and suppose that the pure pixel assumption holds.
Also, for illustration simplicity,
assume without loss of generality (w.l.o.g.) that the first $N$ indices of the measured pixels are pure pixel indices, with index $k$ being a pure pixel index of endmember $k$ for $k=1,\ldots,N$.
Following the discussion in Section~\ref{sec:problem_setup},
we have ${\bf X}_{1:N} = [~ {\bf a}_1, \ldots, {\bf a}_N ~] = {\bf A}$ (in the noiseless case).
Subsequently, we can perform the representation ${\bf X} = {\bf X}{\bf C}$ by setting
\begin{equation} \label{eq:C_exa}
{\bf C}= \begin{bmatrix} {\bf s}[1] & {\bf s}[2] & \ldots & {\bf s}[L] \\
                            {\bf 0} & {\bf 0} & \ldots & {\bf 0}
\end{bmatrix}.
\end{equation}
Note that \eqref{eq:C_exa} automatically satisfies the constraints ${\bf C} \geq {\bf 0}$ and ${\bf 1}^T {\bf C}= {\bf 1}^T$, owing to the non-negative and sum-to-one natures of the abundance vectors ${\bf s}[n]$.
Thus, \eqref{eq:C_exa} is a feasible point of Problem~\eqref{eq:SD-MMV}.
Moreover, by letting $\Lambda= {\rm rowsupp}({\bf C})$ for \eqref{eq:C_exa} (which is simply $\Lambda = \{ 1, 2, \ldots, N \}$ in this example),
we observe that ${\bf X}_\Lambda = {\bf A}$, that is, the true endmember signature matrix.
Note that for cases where pure pixel indices are arbitrarily placed,
the same argument holds;
see \cite{Elhamifar2012}, particularly (12) there.
The intuition of the above observation is that {\em by solving the SD-MMV problem~\eqref{eq:SD-MMV},
we may recover a complete pure pixel index set $\Lambda$, and consequently, identify the true endmember signature matrix.}
A rigorous analysis confirming the validity of this intuition, with noise also being taken into account, will be shown later (Theorem~\ref{thm:SD-MMV-noisy}).
It is worthwhile to point out that the SD-MMV formulation in Problem~\eqref{eq:SD-MMV} does not assume knowledge of 
the model order $N$.
Hence, fundamentally the SD-MMV formulation also provides us with the opportunity to identify
the model order,
as indicated by $\| {\bf C} \|_{\rm row-0}$.

As is well known in compressive sensing,
a sparse MMV problem can be tackled either by convex relaxations~\cite{Tropp2006,Eldar2010average}, which replaces $\| {\bf C} \|_{\rm row-0}$ with a convex function, or by greedy pursuit~\cite{Chen2006,Tropp2006pt2}, which employs simple atom selection schemes.
In the specific context of the SD-MMV model, the existing studies mainly focus on convex relaxations~\cite{Esser2012,Elhamifar2012}.
It is shown that a mixed $\ell_q$-$\ell_1$-norm relaxation of Problem~\eqref{eq:SD-MMV}, with $q > 1$,  can guarantee exact recovery of a desired ${\bf C}$ (such as \eqref{eq:C_exa} in the above example) in the noiseless case and under the pure pixel assumption.
This exact recovery result is meaningful,
and different from results seen in the MMV model.
In general, exact recovery results for the sparse MMV model usually require certain conditions on the dictionary ${\bf B}$~\cite{Chen2006},
such as the mutual coherence level of ${\bf B}$.
In SD-MMV, it turns out that the key behind achieving exact recovery is the pure pixel assumption, rather than the mutual coherence or similar measures.
However, there is an arguably restrictive assumption behind the previous convex relaxation works---that one endmember cannot have
repeated pure pixels.

In the remaining part of this paper, the greedy approach to SD-MMV will be considered.

\section{Greedy Algorithm for SD-MMV}
\label{sec:SD-SOMP}

In this section, we consider a greedy algorithm for the SD-MMV formulation.
The exact recovery condition of the algorithm in the noiseless case will be analyzed,
and the relationship to some existing HU algorithms revealed.


Specifically, we concentrate on
simultaneous orthogonal matching pursuit (SOMP)~\cite{Tropp2006pt2,Chen2006},
a well-known greedy algorithm for the sparse MMV problem.
SOMP approximates the MMV problem \eqref{eq:MMV} by selecting one atom at a time.
To describe it,
suppose that we have previously selected a number of $k-1$ atoms,
represented by the index set $\Lambda_{k-1} = \{ \hat{n}_1, \ldots, \hat{n}_{k-1} \} \subset \{ 1, \ldots, K \}$.
In particular, ${\bf B}_{\Lambda_{k-1}}= [~ {\bf b}_{\hat{n}_1}, \ldots, {\bf b}_{\hat{n}_{k-1}} ~]$ contains the previously selected atoms.
To select a new atom, SOMP first forms a residual
\begin{equation} \label{eq:R}
{\bf R}_{k-1} = {\bf P}^\perp_{{\bf B}_{\Lambda_{k-1}}} {\bf X},
\end{equation}
where ${\bf P}^\perp_{{\bf B}_{\Lambda_{k-1}}}$ is the orthogonal complement projector of ${\bf B}_{\Lambda_{k-1}}$ and
is used to remove components of ${\bf B}_{\Lambda_{k-1}}$ from ${\bf X}$.
Then, the new atom is chosen by the {\em greedy selection step}
\begin{equation} \label{eq:greed_sel}
\hat{n}_k = \arg \max_{i=1,\ldots,K} \| {\bf R}_{k-1}^T {\bf b}_i \|_q,
\end{equation}
where $q \geq 1$ is a prescribed constant.
The intuition behind \eqref{eq:greed_sel} is that if ${\bf R}_{k-1}$ is mostly contributed by one atom,
then $\| {\bf R}_{k-1}^T {\bf b}_i \|_q$ is likely to be the largest at that atom.
Once $\hat{n}_k$ is obtained, we form the new atom selection index set via $\Lambda_k = \Lambda_{k-1} \cup \{ \hat{n}_k \}$.
The steps in \eqref{eq:R}-\eqref{eq:greed_sel} is repeated (with $k$ increased by one each time) until a stopping rule is satisfied.


\begin{algorithm}[!h]
\SetAlgoNoLine
\SetKwInOut{Input}{input}\SetKwInOut{Output}{output}
\SetKwRepeat{Repeat}{repeat}{until}

\Input{${\bf X}$;}

$k=0$; ${\Lambda}_0=\varnothing$; ${\bf R}_0={\bf X}$;

\Repeat{a stopping rule is satisfied}{

$k=k+1$;

$\displaystyle \hat{n}_k=\arg\max_{n=1,\ldots,L}~ \left\| {\bf R}_{k-1}^T{\bf x}[n]
\right\|_q$;

$\Lambda_k = \Lambda_{k-1}\cup \{\hat{n}_k\}$;

${\bf R}_k = {\bf P}^{\perp}_{{\bf X}_{\Lambda_k}}{\bf X}$;

}

\Output{$\hat{\Lambda}={\Lambda}_k$.}

\caption{$\ell_q$ SD-SOMP}\label{Algo:lq-SD-SOMP}
\end{algorithm}

Now, we apply SOMP to the SD-MMV problem \eqref{eq:SD-MMV} by replacing ${\bf B}$ with ${\bf X}$.
The resulting algorithm, summarized in Algorithm~\ref{Algo:lq-SD-SOMP}, will be called {\em $\ell_q$ SD-SOMP} in the sequel.
We are interested in the fundamental natures of $\ell_q$ SD-SOMP.
First, we show that $\ell_q$ SD-SOMP is provably sound in terms of guarantee of exact recovery.


\begin{Theorem} \label{thm:SD-SOMP}
Consider $\ell_q$ SD-SOMP in the noiseless case and under the pure pixel assumption.
For any $q \in (1, \infty]$, the following results hold.
\begin{itemize}
\item[1.] For  $k \in \{ 1, \ldots, N \}$, the index $\hat{n}_k$ obtained in the greedy selection step (cf. Step 4 in Algorithm~\ref{Algo:lq-SD-SOMP}) is a pure pixel index.
\item [2.] The index set $\Lambda_N = \{ \hat{n}_1, \ldots, \hat{n}_N \}$ is a complete pure pixel index set.
\end{itemize}
\end{Theorem}

The proof of Theorem~\ref{thm:SD-SOMP} is given in Appendix~\ref{sec:proof_SD-SOMP}.
Theorem~\ref{thm:SD-SOMP} asserts that $\ell_q$ SD-SOMP attains exact pure pixel index identifiability in the absence of noise.
In particular, the simple greedy scheme by $\ell_q$ SD-SOMP gives the same exact recovery guarantee as that by the previous convex relaxation works~\cite{Esser2012,Elhamifar2012}.
We should also note that while the convex relaxation works require the assumption of non-repeated pure pixel indices to achieve exact recovery,
$\ell_q$ SD-SOMP has no such restriction.
In the above result, we do not specify how
the model order
$N$ is identified.
In fact, identifying $N$ with greedy pursuit is trivial in the noiseless noise---
it is immediate from Theorem~\ref{thm:SD-SOMP} that if the stopping rule (cf. Step 7 in Algorithm~\ref{Algo:lq-SD-SOMP}) is $\| {\bf R}_k \|_F= 0$, then $\ell_q$ SD-SOMP stops at $k=N$.
It should however be stressed that in the noisy case,
identifying $N$ is nontrivial.
This issue will be addressed in the next section.

Second, there is a connection between $\ell_q$ SD-SOMP and some existing HU algorithms.
Let us consider the special case of $q = \infty$.

\begin{Observation} \label{obs:SD-SOMP}
For $\ell_\infty$ SD-SOMP, the greedy selection step in Line 4 of Algorithm~\ref{Algo:lq-SD-SOMP} can be equivalently written as
\begin{equation}
\hat{n}_k = \arg \max_{n=1, \ldots, L} \left\| {\bf P}^\perp_{ {\bf X}_{\Lambda_{k-1}} } {\bf x}[n]   \right\|_2.
\end{equation}
\end{Observation}

Observation~\ref{obs:SD-SOMP} appears in the proof of Theorem~\ref{thm:SD-SOMP} (See Appendix~\ref{sec:proof_SD-SOMP}).
Interestingly, we observe that if we set the stopping rule as $k  \geq N$ (or run $N$ iterations and then terminates),
then
$\ell_\infty$ SD-SOMP takes exactly the same form as SPA~\cite{MC01}.
We should remark that SPA falls in a family of recursive pure pixel search algorithms, e.g., \cite{MC01,ATGP,GENE},
and that SPA is also known to be very similar to SVMAX~\cite{VMAX} under the Winter simplex volume maximization formulation;
see \cite{Ma2014} for further discussion.
As an aside, SPA appears to be very special---the same algorithm can be derived in three different ways, namely, recursive pure pixel search, simplex volume maximization, and now, greedy SD-MMV pursuit.

Before finishing this section, we should point out that one can also employ other greedy algorithms to process the SD-MMV problem. For example, in \cite{fu2013greedy}, we derive another greedy SD-MMV algorithm using the reduced MMV and boost framework~\cite{Mishali2008}.
Interestingly, the resulting algorithm turns out to be similar to vertex component analysis (VCA)~\cite{VCA}, a popular pure pixel search algorithm.
We skip this result due to the limit of space.

\section{Sparse SD-MMV Model in the Noisy Case}

In this section, we consider the sparse SD-MMV model and its greedy algorithm in the case where noise is present and the number of endmembers is unknown.

\subsection{A Robust SD-MMV Formulation}

In the noisy case, we propose to adopt the following robust SD-MMV formulation
\begin{subequations} \label{eq:SD-MMV-noisy}
\begin{align}
\min_{ {\bf C} \in \mathbb{R}^{L \times L} } & ~ \| {\bf C} \|_{\rm row-0}    \\
{\rm s.t.}       & ~ \| {\bf x}[n]- {\bf X}{\bf c}_n \|_2 \leq \delta, ~ n=1,\ldots,L,  \label{eq:SD-MMV-noisy_b} \\
                 & ~ {\bf C} \geq {\bf 0}, {\bf 1}^T {\bf C} = {\bf 1}^T, \label{eq:SD-MMV-noisy_c}
\end{align}
\end{subequations}
where $\delta \geq 0$ is a prespecified constant.
Problem~\eqref{eq:SD-MMV-noisy} is reminiscent of the basis pursuit denoising formulation in compressive sensing,
where the former seeks to approximate ${\bf X} = {\bf X}{\bf C}$ within a certain tolerance level (specified by $\delta$) while minimizing the number of atoms involved at the same time.
Also, Problem~\eqref{eq:SD-MMV-noisy} reduces to the noiseless SD-MMV formulation~\eqref{eq:SD-MMV} when $\delta = 0$.
Let ${\bf C}_{\rm opt}$ be an optimal solution to Problem~\eqref{eq:SD-MMV-noisy}.
Following the SD-MMV concept described in Section~\ref{sec:SD-MMV},
we estimate a complete pure pixel index set by using $\Lambda_{\rm opt} = {\rm rowsupp}({\bf C}_{\rm opt})$.
In particular, $|\Lambda_{\rm opt}|$ (or $\| {\bf C}_{\rm opt} \|_{\rm row-0})$ provides an estimate of
the model order
$N$, and ${\bf X}_{\Lambda_{\rm opt}}$ an estimate of the endmember signature matrix ${\bf A}$.

Before proceeding to the development of greedy pursuit for Problem~\eqref{eq:SD-MMV-noisy},
it is interesting to analyze how robust Problem~\eqref{eq:SD-MMV-noisy} can be in the noisy case.
Consider the following setting.

\begin{Assumption} \label{assume:1}
The pure pixel assumption holds, and every noise vector ${\bf v}[n]$, $n=1,\ldots,L$, satisfies $\| {\bf v}[n] \|_2 \leq \epsilon$ for some $\epsilon > 0$.
\end{Assumption}

Our robustness analysis is based on a worst-case approach with respect to (w.r.t.) noise.
First, we show two lemmas concerning the model order estimate.
The first lemma is as follows.

\begin{Lemma} \label{lem:1}
Suppose that Assumption~\ref{assume:1} holds.
If $\delta \geq 2 \epsilon$, then there exists a feasible point ${\bf C}$ of Problem~\eqref{eq:SD-MMV-noisy} such that $\| {\bf C} \|_{\rm row-0}=N$.
\end{Lemma}

The proof of Lemma~\ref{lem:1} can be found in Appendix~\ref{app:lem1}.
Lemma~\ref{lem:1} suggests that we can guarantee
\[ \| {\bf C}_{\rm opt} \|_{\rm row-0} \leq N \]
by setting $\delta \geq 2 \epsilon$;
or, in words, an optimal solution to Problem~\eqref{eq:SD-MMV-noisy} does not overestimate the number of endmembers $N$ for a sufficiently large $\delta$.
The second lemma is described below.

\begin{Lemma} \label{lem:2}
Suppose that Assumption~\ref{assume:1} holds.
Any feasible point ${\bf C}$ of Problem~\eqref{eq:SD-MMV-noisy} must satisfy the following property: For each $k \in \{1,\ldots,N \}$, there exists an index $\hat{n}_k \in {\rm rowsupp}({\bf C})$ such that
\begin{equation} \label{eq:lem:2:bound}
\| {\bf e}_k - {\bf s}[\hat{n}_k]  \|_1 \leq \frac{2(\delta + 2\epsilon)}{\sigma_{\rm min}({\bf A})},
\end{equation}
where $\sigma_{\rm min}({\bf A})$ is the smallest singular value of ${\bf A}$.
In addition, if the condition
\begin{equation} \label{eq:lem:2:bound2}
\frac{2(\delta + 2 \epsilon)}{\sigma_{\rm min}({\bf A})} < 1
\end{equation}
holds, then each $\hat{n}_k$ is distinct; i.e., $\hat{n}_k \neq \hat{n}_j$ for all $k \neq j$, $k, j \in \{ 1,\ldots, N \}$.
\end{Lemma}
The proof of Lemma~\ref{lem:2} can be found in Appendix~\ref{app:lemma2}.
Lemma~\ref{lem:2} implies that
if $\delta$ is chosen such that \eqref{eq:lem:2:bound2} holds, then
\[ \| {\bf C}_{\rm opt} \|_{\rm row-0} \geq N; \]
that is, an optimal solution to Problem~\eqref{eq:SD-MMV-noisy} does not underestimate $N$ for a sufficiently small $\delta$.
This, together with the non-overestimating implication of Lemma~\ref{lem:1}, are vital---by satisfying both the conditions in Lemmas~\ref{lem:1}-\ref{lem:2}, we can achieve $\| {\bf C}_{\rm opt} \|_{\rm row-0} = N$.
With this in mind, we prove in Appendix~\ref{app:SD-MMV-noisy} the following claim.

\begin{Theorem} \label{thm:SD-MMV-noisy}
Consider the SD-MMV problem~\eqref{eq:SD-MMV-noisy} under Assumption~\ref{assume:1}.
Suppose that
\begin{equation}
\epsilon < \frac{\sigma_{\rm min}({\bf A}) \cdot \min\{ 1, d({\bf S})\}}{8},
\label{eq:exact_reov}
\end{equation}
where
\begin{equation} \label{eq:dS}
d({\bf S}) = \min_{k=1,\ldots,N} \min_{\substack{n=1,\ldots,L, \\ {\bf s}[n] \neq {\bf e}_k }}
    \| {\bf e}_k - {\bf s}[n] \|_1
\end{equation}
is a distance measure between pure and non-pure abundance pixels.
Then, for any $\delta \in [2 \epsilon, \sigma_{\rm min}({\bf A})/2 - 2 \epsilon)$,
an optimal solution ${\bf C}_{\rm opt}$ to Problem~\eqref{eq:SD-MMV-noisy} exactly recovers a complete pure pixel index set;
specifically, $\Lambda_{\rm opt} = {\rm rowsupp}({\bf C}_{\rm opt})$ is a complete pure pixel index set.
\end{Theorem}

Theorem~\ref{thm:SD-MMV-noisy} implies that for a sufficiently small noise level,
the SD-MMV problem~\eqref{eq:SD-MMV-noisy} is robust against noise perturbations.
Also, as an immediate corollary of Theorem~\ref{thm:SD-MMV-noisy},
the noiseless SD-MMV formulation~\eqref{eq:SD-MMV} guarantees exact recovery in the noiseless case (set $\delta= \epsilon= 0$).
We should note that the exact recovery condition in \eqref{eq:exact_reov} is a provable bound, taking care of the worst possible noise under Assumption~\ref{assume:1}.
In practice, one may expect a
much better
noise tolerance than that in \eqref{eq:exact_reov}.
Furthermore, Theorem~\ref{thm:SD-MMV-noisy} suggests how $\delta$ should be chosen; specifically, it suffices to choose $\delta= 2 \epsilon$.
Two additional technical remarks are in order.




\begin{Remark} \label{rem:dS}
We should discuss the distance measure $d({\bf S})$, which appears in \eqref{eq:exact_reov} as a performance limiting factor.
The measure $d({\bf S})$ describes the proximity of pure and non-pure pixels in a smallest possible sense, and it can be small when there are near-pure pixels (or non-pure pixels that are close to pure pixels).
Thus, at first look of \eqref{eq:exact_reov}, it seems that achieving exact recovery requires a very small noise level.
This is not exactly true:
near-pure pixels can also be seen as noise-perturbed pure pixels.
By remodeling near-pure pixels as pure pixels\footnote{Specifically, we can write ${\bf x}[n] = {\bf a}_i + \tilde{\bf v}[n]$, $\tilde{\bf v}[n] = {\bf a}_i (s_i[n]-1) +\sum_{j \neq i} {\bf a}_j s_j[n] + {\bf v}[n]$, for some $i$, and see $\tilde{\bf v}[n]$ as noise satisfying Assumption~\ref{assume:1}.},
the value of $d({\bf S})$ can be improved, although such remodeling also comes with a price of increasing the noise level.
\end{Remark}


\begin{Remark}
Following the discussion in the previous remark, one may wonder whether good recovery results can be established without $d({\bf S})$.
The answer is yes.
\begin{Corollary} \label{cor:SD-MMV-noisy}
Consider the SD-MMV problem~\eqref{eq:SD-MMV-noisy} under Assumption~\ref{assume:1}.
Suppose that
\begin{equation}
\epsilon < \frac{\sigma_{\rm min}({\bf A})}{8}.
\end{equation}
Then, for any $\delta \in [2 \epsilon, \sigma_{\rm min}({\bf A})/2 - 2 \epsilon)$,
an optimal solution ${\bf C}_{\rm opt}$ to Problem~\eqref{eq:SD-MMV-noisy} satisfies $\| {\bf C} \|_{\rm row-0} = N$.
Also, by denoting ${\rm rowsupp}({\bf C}_{\rm opt}) = \{ \hat{n}_1, \ldots, \hat{n}_N \}$, the endmember estimate $\{ {\bf x}[\hat{n}_1],\ldots, {\bf x}[\hat{n}_N] \}$ satisfies
\[ \| {\bf a}_k - {\bf x}[\hat{n}_k] \|_2 \leq 2(\delta+2\epsilon)  \frac{\displaystyle \max_{i=1,\ldots,N} \| {\bf a}_i \|_2}{\sigma_{\rm min}({\bf A})} + \epsilon. \]
\end{Corollary}


Corollary \ref{cor:SD-MMV-noisy} is a side-product of the proof of Theorem~\ref{thm:SD-MMV-noisy}; see Appendix \ref{app:cor-SD-MMV-noisy} for the proof.
Corollary \ref{cor:SD-MMV-noisy} indicates that even if Problem~\eqref{eq:SD-MMV-noisy} may not guarantee exact recovery in a complete sense,
it can still guarantee exact recovery of the true model order under a more relaxed condition on the noise level.
It is also worthwhile to note that Corollary~\ref{cor:SD-MMV-noisy} pins down a provable bound on the endmember estimation accuracy,
again, under a more relaxed noise setting.
\end{Remark}



\subsection{Greedy Pursuit for the Robust SD-MMV Formulation}

We now turn our attention back to greedy pursuit.
Our development in the noisy case is the same as before---use a greedy method, specifically, $\ell_q$ SD-SOMP in Section~\ref{sec:SD-SOMP}, to iteratively select atoms.
What requires design here is that of the stopping rule, which, as discussed, is trivial in the noiseless case but not as obvious in the noisy case.
Moreover, we hope that with a proper stopping rule design, the resulting greedy algorithm can have some performance guarantee that links up with that of the robust SD-MMV formulation in the last subsection.
Our study starts with the following stopping rule:

\vspace{\baselineskip}
\noindent
\fbox{
\parbox{.95\linewidth}{
{\bf Stopping Rule~1:}
Stop if
there exists a feasible point ${\bf C}$ of Problem~\eqref{eq:SD-MMV-noisy} whose row support is fixed as ${\rm rowsupp}({\bf C}) = \Lambda_k$.
}}
\vspace{\baselineskip}

The idea of Stopping Rule 1 is natural:
progressively increase the estimated model order 
until
we can find a feasible solution to Problem~\eqref{eq:SD-MMV-noisy}.
To implement Stopping Rule 1, first note that
for any ${\bf C}$ satisfying ${\rm rowsupp}({\bf C})= \Lambda_k$, we can represent
\[ \| {\bf x}[n] - {\bf X}{\bf c}_n \|_ 2 = \| {\bf x}[n]- {\bf X}_{\Lambda_k} \bar{\bf c}_n \|_2 \]
where $\bar{\bf c}_n \in \mathbb{R}^{|\Lambda_k|}$ denotes a subvector of ${\bf c}_n$, obtained by choosing the elements of ${\bf c}_n$ whose row indices lie in $\Lambda_k$.
By solving the following fully constrained least squares (FCLS) problems
\begin{equation} \label{eq:e_n}
e_n = \min_{ \bar{\bf c} \geq {\bf 0}, {\bf 1}^T \bar{\bf c}= 1 }  \| {\bf x}[n]- {\bf X}_{\Lambda_k} \bar{\bf c} \|_2^2
\end{equation}
for all $n=1,\ldots,L$, which can be readily done by applying an available solver,
we can verify the existence of the desired feasible ${\bf C}$ by checking whether $e_n \leq \delta^2$ holds  for all $n$.

Now, we come to the following question---can $\ell_q$ SD-SOMP guarantee exact recovery of a complete pure pixel index set?
To address this analysis problem, we first need a key result by Gillis and Vavasis.

\begin{Fact}[Theorem 3, \cite{Gillis2012}] \label{fact:SPA}
Consider SPA, or equivalently, running $\ell_\infty$ SD-SOMP for $N$ iterations.
Suppose that Assumption~\ref{assume:1} holds.
If
\begin{equation}  \label{eq:SPA_eps_bnd}
\epsilon < \frac{ \sigma_{\rm min}({\bf A}) }{ 4 \sqrt{N} \eta({\bf A}) },
\end{equation}
where $\eta({\bf A}) \geq 1$ is a constant that is proportional to
\[ \eta({\bf A}) \leq \mathcal{O}\left(  \frac{ \max_{i=1,\ldots,N} \| {\bf a}_i \|_2^2 }{  \sigma_{\rm min}^2({\bf A})  } \right),
\]
then the pure pixel index estimates $\hat{n}_1,\ldots,\hat{n}_N$ satisfy an error bound
\begin{equation} \label{eq:SPA_e_bnd}
\| {\bf a}_{\pi_k} - {\bf x}[\hat{n}_k] \|_2 \leq \epsilon \cdot \eta({\bf A})
\end{equation}
for some permutation $\bm \pi$ of $\{ 1, \ldots, N \}$.
\end{Fact}

We should mention that the Gillis-Vavasis theorem described above is a worst-case provable result under the scenario of known model order.
Using Fact~\ref{fact:SPA} and the proof techniques in Lemmas~\ref{lem:1}-\ref{lem:2}, we establish
an
exact recovery condition as follows.

\begin{Theorem}  \label{thm:SOMP1}
Consider $\ell_\infty$ SD-SOMP with Stopping Rule 1.
Suppose that Assumption~\ref{assume:1} holds, and that
\begin{equation} \label{eq:SOMP1_e_bnd}
\epsilon < \frac{ \sigma_{\rm min}({\bf A}) \cdot \min\{ 1, d({\bf S}) \} }{ 4 \sqrt{N} \eta({\bf A}) },
\end{equation}
where $d({\bf S})$ is defined in Theorem~\ref{thm:SD-MMV-noisy}.
Then, for any $\delta \in [ 2\epsilon, \sigma_{\rm min}({\bf A})- 2\epsilon)$,
$\ell_\infty$ SD-SOMP guarantees exact recovery in the sense that
the output $\hat{\Lambda}$ is a complete pure pixel index set.
\end{Theorem}

The proof of Theorem~\ref{thm:SOMP1} is relegated to Appendix \ref{app:thm-SOMP1}.
It is interesting to compare Theorems~\ref{thm:SD-MMV-noisy} and \ref{thm:SOMP1}---their exact recovery conditions differ only by a factor of $0.5 \sqrt{N} \eta({\bf A})$, which can be interpreted as a performance loss factor owing to the use of
greedy approximation for
the SD-MMV problem~\eqref{eq:SD-MMV-noisy}.

While
we have shown that $\ell_\infty$ SD-SOMP with Stopping Rule 1
exhibits exact recovery guarantee in the noisy case,
it has a drawback.
Specifically, Stopping Rule 1 requires solving a number of $L$ FCLS problems (cf. Problem~\eqref{eq:e_n}) at each greedy iteration, which can be computationally intense in practice.
On the other hand, while the proof of Theorem~\ref{thm:SOMP1} is for Stopping Rule 1, it reveals a hint on designing a simpler stopping rule that attains the same exact recovery guarantee as Stopping Rule~1.
The result is as follows.

\begin{Theorem}  \label{thm:SOMP2}
$\ell_\infty$ SD-SOMP with the following stopping rule

\vspace{\baselineskip}
\noindent
\fbox{
\parbox{.95\linewidth}{
{\bf Stopping Rule~2:}
{\rm Stop if}
\begin{equation} \label{eq:prob_rule2}
\min_{ \bar{\bf c}  \geq {\bf 0}, {\bf 1}^T\bar{\bf c} = 1 }
   \| {\bf x}[\hat{n}_{k+1}] - {\bf X}_{\Lambda_k} \bar{\bf c} \|_2
   \leq \delta,
\end{equation}
{\rm where $\hat{n}_{k+1}$ is the index obtained by the greedy selection step at the next iteration.
}}}
\vspace{\baselineskip}

\noindent
attains the same exact recovery condition as in Theorem~\ref{thm:SOMP1}.
\end{Theorem}

The proof of Theorem~\ref{thm:SOMP2} is described in Appendix \ref{app:thm-SOMP2}.
Note that Stopping Rule 2 requires solving only one FCLS problem at each greedy iteration, which is much less than that in Stopping Rule 1.
It is interesting to mention that similar forms of Stopping Rule 2 have been considered in some prior works \cite{GENE,ATGP-NPD}.
The most similar one is GENE-CH~\cite{GENE},
wherein the authors proposed a Neyman-Pearson detection rule for \eqref{eq:prob_rule2} by applying a rather strong model.
In particular, the optimal solution to the problem at the left-hand side (LHS) of \eqref{eq:prob_rule2} is modeled as the true abundance,
which intuitively seems reasonable, but has no strong theoretical basis to support in the noisy case.
In comparison, our analysis is rigorous from a worst-case provable bound perspective;
Assumption~\ref{assume:1} is the only assumption in our proof.
Moreover, GENE-AH~\cite{GENE} is a variant of GENE-CH, wherein the sum-to-one constraint in \eqref{eq:prob_rule2} is relaxed.
Also, an earlier work, ATGP-NPD \cite{ATGP-NPD}, considers the unconstrained least squares version of \eqref{eq:prob_rule2}; it uses the Neyman-Pearson detection rule, which is later adopted in GENE-AH and GENE-CH.
By our analysis experience, both the non-negativity and sum-to-one constraints in \eqref{eq:prob_rule2} are vital in bounding the noise effects.

Our analysis also reveals some more technical results.

%
%
%

\begin{Remark}
Like Corollary~\ref{cor:SD-MMV-noisy}, we
can also
prove a more relaxed condition under which $\ell_\infty$ SD-SOMP can at least guarantee correct identification of the model order.

\begin{Corollary} \label{cor:SOMP}
Suppose that Assumption~\ref{assume:1} holds.
$\ell_\infty$ SD-SOMP with Stopping Rule 1 satisfies $|\hat{\Lambda}|=N$ if
\begin{equation} \label{eq:cor:SOMP:eps_bnd}
\epsilon < \frac{ \sigma_{\rm min}({\bf A}) }{  4( 1 + \sqrt{N} \eta({\bf A}) ) },
\end{equation}
and for $\delta \in [(1+\eta({\bf A})) \epsilon, \sigma_{\rm min}({\bf A})- (3 + \eta({\bf A})) \epsilon )$.
$\ell_\infty$ SD-SOMP with Stopping Rule 2 satisfies $|\hat{\Lambda}|=N$ if
\begin{equation} \label{eq:cor:SOMP:eps_bnd2}
\epsilon < \frac{ \sigma_{\rm min}({\bf A}) }{  4 ( 1.25 + \sqrt{N}  \eta({\bf A}) ) },
\end{equation}
and for $\delta \in [(1+\eta({\bf A})) \epsilon, \sigma_{\rm min}({\bf A})- (4 + 2\eta({\bf A})) \epsilon )$.
In both cases, the endmember estimation error bound in \eqref{eq:SPA_e_bnd} is satisfied.
\end{Corollary}
The proof of Corollary~\ref{cor:SOMP} is relegated to Appendix~\ref{app:cor-SOMP}.
Comparing Corollary~\ref{cor:SOMP} and Fact~\ref{fact:SPA},
we see that the error bounds in \eqref{eq:cor:SOMP:eps_bnd}-\eqref{eq:cor:SOMP:eps_bnd2} are only slightly worse than \eqref{eq:SPA_eps_bnd}; which means that pure pixel identification without knowing the model order does not incur a significant performance loss than that knowing the model order.
\end{Remark}
\begin{Remark} \label{rem:lq_SD_SOMP}
Although we focused only on $\ell_\infty$ SD-SOMP in this section, all the above results can be readily extended to $\ell_q$ SD-SOMP for any
$q \in (1, \infty)$. The key insight is to apply the Gillis-Vavasis theorem again on the function $f({\bf x})=\|{\bf X}^T{\bf x}\|_q$; see details in \cite{Gillis2012}.
It can be shown that
Fact~\ref{fact:SPA} still holds, except that $\eta({\bf A})$ should be replaced by
\[\eta({\bf A})\leq {\cal O}\left( \alpha_q \cdot {\kappa}({\bf X})\cdot \frac{\max_{i=1,\ldots,N}{\|{\bf a}_i\|_2^2}}{\sigma_{\min}^2({\bf A})} \right), \]
where $\alpha_q \geq 1$ is a constant related to the (local) Lipschitz constant of the gradient of $\| {\bf x} \|_q$ and the (local) strong convexity parameter of $\| {\bf x} \|_q$,
and ${\kappa}({\bf X}) = \sigma_{\rm max}({\bf X})/\sigma_{\rm min}({\bf X}) \geq 1$ is the condition number of ${\bf X}$.
Comparing the above equation and the $\eta({\bf A})$ in Fact~\ref{fact:SPA},
we see no advantage of using $q \in (1, \infty)$ over $q= \infty$ from a provable bound viewpoint.
Before we close this remark, we should mention the concurrent work \cite{gillis2014enhancing};
roughly speaking, it shows that $\eta({\bf A})$ can be improved via preconditioning methods.
\end{Remark}

\section{Simulations}
\subsection{Simulation Settings}
In this section,
synthetic hyperspectral images are used to demonstrate the performance of $\ell_q$ SD-SOMP.
The endmembers are picked from a subset of the U.S. Geological Survey (U.S.G.S.) library~\cite{USGS2007}, and they are minerals such as
Carnallite,
Ammonioalunite,
Biotite,
Actinolite,
Almandine,
Ammonio-jarosite,
Andradite,
Antigorite,
Axinite,
Brucite,
Carnallite,
Chlorite,
Clinochlore,
Clintonite,
Corundum,
Diaspore,
Elbaite,
Erionite$+$Merlinoit,
Galena,
Goethite,
and Halloysite.
The number of frequency bands is $M=224$.
The abundance vectors are randomly generated following the uniform Dirichlet distribution.
Pure pixels are manually added.
The noise is zero-mean white Gaussian, both spectrally and spatially.
The SNR is defined as
${\rm SNR}=\frac{\sum_{n=1}^L\|{\bf A}{\bf s}[n]\|_2^2}{ML\sigma^2}$,
where $\sigma^2$ is the noise power.
In all the simulations,
the model order
$N$ is unknown to the HU algorithms.

The settings for $\ell_q$ SD-SOMP are as follows.
Stopping Rule 2 is employed.
Following the design guideline suggested by Theorem~\ref{thm:SOMP2}, we set $\delta=2{\epsilon}$.
The noise bound $\epsilon$ is unknown in practice,
and we estimate it by using
multiple linear regression~\cite{Hysime} to obtain gross estimates of the noise vectors ${\bf v}[n]$,
and then using the estimated noise vectors to determine $\epsilon$.
By our empirical experience, the above procedure is able to provide a reasonable noise bound.

We choose GENE-CH and GENE-AH \cite{GENE}  as benchmarks.
Regarding their settings, we should mention that GENE-CH/AH employs a preprocessing step for noise suppression.
Specifically, the preprocessing step projects the measured pixel vectors into an ${N}_{\max}$-dimensional ($N\leq N_{\max}\ll M$) subspace, where ${N}_{\max}$ is the largest possible number of endmembers (or a conservative rough guess of $N$).
The projection step is referred to as \emph{inexact affine set fitting-based dimension reduction} (ASF-DR); see \cite{GENE} for details.
We set $N_{\max}=50$ for GENE-CH/AH in our simulations.
Also, unless specified, the false alarm probability specification of GENE-CH/AH is set to ${P}_{\rm FA}=10^{-6}$.

We use the detection probability
\[{\rm Prob}\{\hat{\Lambda}=\Lambda\},\]
as our performance measure, where $\Lambda$ is the (true) complete set of pure pixel indices,
and $\hat{\Lambda}$ denotes an estimate of $\Lambda$ from an HU algorithm.
Note that a high detection probability indicates not only accurate estimation of pure pixel indices, but also accurate estimation of the model order $N$.
The number of trials for evaluating the detection probability is 100.

\subsection{Simulation Results}

\begin{figure}
	\centering
		\includegraphics[width=.45\textwidth]{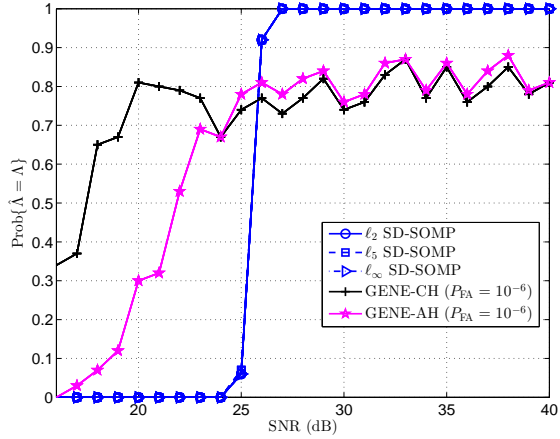}
	\caption{Detection probability performance w.r.t. the SNRs. $(N,L)=(10,5000)$.}
	\label{fig:NEDE_SNR}
\end{figure}

Fig.~\ref{fig:NEDE_SNR} shows the detection probabilities of $\ell_q$ SD-SOMP and GENE-CH/AH under various SNRs.
The true number of endmembers is $N= 10$, and the number of pixels $L= 5,000$.
One can see that when SNR$\leq 26$dB, GENE-CH exhibits the highest detection probabilities.
When SNR$\geq 26$dB, the detection probabilities of $\ell_q$ SD-SOMP (with $q=2,5,\infty$) sharply rise to one.
This observation is consistent with our analysis: $\ell_q$ SD-SOMP is robust to noise under sufficiently high SNRs.
Moreover, we see that $\ell_q$ SD-SOMP for $q=2,5$ essentially yields the same performance as $\ell_\infty$ SD-SOMP.
Since $\ell_\infty$ SD-SOMP is easier to implement and $\ell_q$ SD-SOMP for $q \in (1, \infty)$ does not exhibit a better provable robustness than $\ell_\infty$ SD-SOMP  (see Remark~\ref{rem:lq_SD_SOMP}),
we will consider only $\ell_\infty$ SD-SOMP hereafter.

\begin{figure}
	\centering
		\includegraphics[width=.45\textwidth]{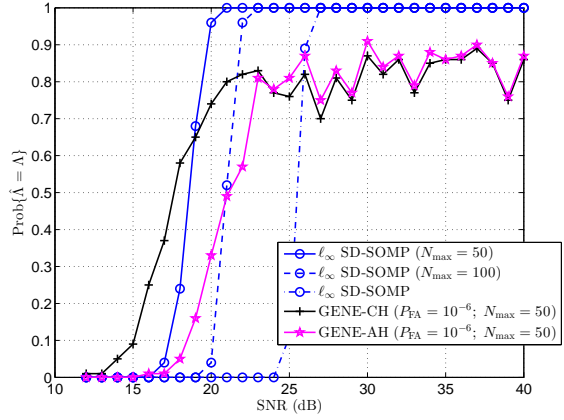}
	\caption{Performance under various $N_{\rm max}$. $(N,L)=(10,5000)$.}
	\label{fig:DR_Nmax}
\end{figure}

Our development has been based on a plain hyperspectral model
(i.e., \eqref{eq:model}), which does not consider any preprocessing for noise suppression.
In fact, we can do the latter and performance can be improved.
Fig.~\ref{fig:DR_Nmax} shows the performance of $\ell_\infty$ SD-SOMP when the inexact ASF-DR step is employed.
We see that the detection probability of $\ell_\infty$ SD-SOMP with ASF-DR is considerably enhanced compared to that without ASF-DR.

\begin{figure}
	\centering
		\includegraphics[width=.45\textwidth]{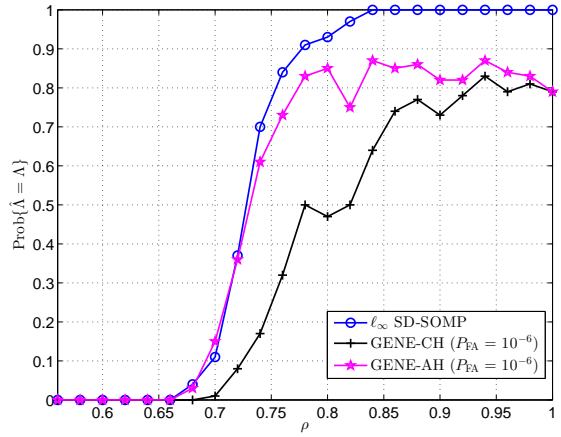}
	\caption{Detection probability performance under no pure pixels. $(N,L)=(10,5000)$; SNR$=35$dB.}
	\label{fig:NEDE_rho}
\end{figure}

It is also interesting to examine how $\ell_\infty$ SD-SOMP or other HU algorithms perform when no pure pixel exists.
To carry out such a simulation,
let us define
\[\rho_k = \max_{n=1,\ldots,L}~s_k[n]\]
to be the pure pixel level of endmember $k$.
Since there is no pure pixel, we evaluate another detection probability, ${\rm Prob}\{\hat{\Lambda}=\tilde{\Lambda}\}$,
where $\tilde{\Lambda}=\left\{\tilde{n}_1,\ldots,\tilde{n}_N\right\}$ is an index set of `nearest' pure pixels,
and those nearest pure pixels are picked by
\[\tilde{n}_k = \arg\min_{n=1,\ldots,L}~\|{\bf A}{\bf s}[n]-{\bf a}_k\|_2,\quad \forall k. \]
Fig.~\ref{fig:NEDE_rho} shows the performance of $\ell_\infty$ SD-SOMP under various pure pixel levels,
wherein we set $\rho_k=\rho$ for $k=1,\ldots,N$.
We see that when $\rho\geq 0.85$, $\ell_\infty$ SD-SOMP can always exactly identify $\tilde{\Lambda}$.
Such robustness to violation of the pure pixel assumption can be explained by the robustness of $\ell_q$ SD-SOMP to noise,
since pixels that are close to the true endmembers can be considered as pure pixels corrupted by small noise; see the discussion in Remark~\ref{rem:dS}.

Table~\ref{tab:MOS_Nvary} shows the accuracies of estimating the model order $N$ by various algorithms.
This time, we consider not only $\ell_\infty$ SD-SOMP and GENE-CH/-AH, but also other state-of-the-art methods for estimating the model order; they are HYSIME \cite{Hysime}, HFC and its noise-whitened version, i.e., NWHFC \cite{HFC}, and ATGP-NPD \cite{ATGP-NPD}.
It is seen that in general, both $\ell_\infty$ SD-SOMP and HYSIME give very reliable estimations of $N$.
In particular, $\ell_\infty$ SD-SOMP always achieves correct identification as $N$ varies from $4$ to $16$,
while HYSIME tends to underestimate $N$ when $N=16$ and $20$.

%

%

\ifconfver

\begin{table}[htbp]
  \centering
  \caption{Means$\pm$standard deviations of the estimated numbers of endmembers by various algorithms. SNR$=35$dB$; L=5000$.}
    \resizebox{8.8cm}{!}{ \huge
\begin{tabular}{c|c|c|c|c|c|c}
\hline
\hline
\multirow{2}[4]{*}{Algorithm} & \multirow{2}[4]{*}{$P_{\rm FA}$} & \multicolumn{5}{c}{$N$} \\
\cline{3-7}      &       & 4     & 8     & 12    & 16    & 20 \\
\hline
\hline

$\ell_\infty$ SD-SOMP &       & 4$\pm$0 & 8$\pm$0 & 12$\pm$0 & 16$\pm$0 & 20$\pm$0.197 \\
\hline
HYSIME &       & 4$\pm$0 & 8$\pm$0 & 12$\pm$0 & 15$\pm$0.171 & 17$\pm$0.1 \\
\hline
\multirow{3}[6]{*}{GENE-CH} & $10^{-4}$ & 6.28$\pm$1.89 & 9.19$\pm$1.27 & 13.1$\pm$1.14 & 16.6$\pm$0.861 & 20.2$\pm$0.443 \\
\cline{2-7}      & $10^{-5}$ & 5.08$\pm$1.04 & 8.44$\pm$0.671 & 12.4$\pm$0.715 & 16.1$\pm$0.327 & 20$\pm$0.245 \\
\cline{2-7}      & $10^{-6}$ & 4.39$\pm$0.634 & 8.12$\pm$0.327 & 12.1$\pm$0.239 & 16$\pm$0.1 & 20$\pm$0.174 \\
\hline
\multirow{3}[6]{*}{GENE-AH} & $10^{-4}$ & 5.65$\pm$1.06 & 8.99$\pm$1.07 & 12.8$\pm$0.94 & 16.4$\pm$0.609 & 20$\pm$0.284 \\
\cline{2-7}      & $10^{-5}$ & 4.95$\pm$0.914 & 8.41$\pm$0.653 & 12.3$\pm$0.525 & 16.1$\pm$0.256 & 19.9$\pm$0.356 \\
\cline{2-7}      & $10^{-6}$ & 4.35$\pm$0.575 & 8.1$\pm$0.302 & 12$\pm$0.141 & 16$\pm$0 & 19.7$\pm$0.482 \\
\hline
\multirow{3}[6]{*}{HFC} & $10^{-4}$ & 4$\pm$0 & 5.93$\pm$0.293 & 9$\pm$0 & 8.14$\pm$0.817 & 3.01$\pm$0.1 \\
\cline{2-7}      & $10^{-5}$ & 4$\pm$0 & 5.82$\pm$0.386 & 9$\pm$0 & 7.24$\pm$0.495 & 2.99$\pm$0.1 \\
\cline{2-7}      & $10^{-6}$ & 4$\pm$0 & 5.93$\pm$0.541 & 9$\pm$0.239 & 8.14$\pm$0.284 & 3.01$\pm$0.171 \\
\hline
\multirow{3}[6]{*}{NWHFC} & $10^{-4}$ & 4$\pm$0 & 4.13$\pm$0.367 & 9.36$\pm$0.482 & 7.15$\pm$0.359 & 3.11$\pm$0.314 \\
\cline{2-7}      & $10^{-5}$ & 4$\pm$0 & 4$\pm$0 & 9.08$\pm$0.273 & 7.01$\pm$0.1 & 3.01$\pm$0.1 \\
\cline{2-7}      & $10^{-6}$ & 4$\pm$0 & 4$\pm$0 & 9$\pm$0 & 6.98$\pm$0.141 & 2.99$\pm$0.1 \\
\hline
\multirow{3}[6]{*}{ATGP-NPD} & $10^{-4}$ & 18.4$\pm$1.24 & 24.1$\pm$1.62 & 23.7$\pm$1.49 & 20.9$\pm$1.51 & 20.9$\pm$0.81 \\
\cline{2-7}      & $10^{-5}$ & 15.9$\pm$1.11 & 20.6$\pm$1.27 & 20.6$\pm$1.33 & 18.5$\pm$1.07 & 20.2$\pm$0.647 \\
\cline{2-7}      & $10^{-6}$ & 14.1$\pm$0.886 & 18.5$\pm$1.04 & 18$\pm$1.4 & 17.6$\pm$0.777 & 19.8$\pm$0.512 \\
\hline
\hline
\end{tabular}%
}
  \label{tab:MOS_Nvary}%
\end{table}%

\else

\begin{table}[htbp]
  \centering
  \caption{Means$\pm$standard deviations of the estimated numbers of endmembers by various algorithms. SNR$=35$dB$; L=5000$.}
\begin{tabular}{c|c|c|c|c|c|c}
\hline
\hline
\multirow{2}[4]{*}{Algorithm} & \multirow{2}[4]{*}{$P_{\rm FA}$} & \multicolumn{5}{c}{$N$} \\
\cline{3-7}      &       & 4     & 8     & 12    & 16    & 20 \\
\hline
\hline

$\ell_\infty$ SD-SOMP &       & 4$\pm$0 & 8$\pm$0 & 12$\pm$0 & 16$\pm$0 & 20$\pm$0.197 \\
\hline
HYSIME &       & 4$\pm$0 & 8$\pm$0 & 12$\pm$0 & 15$\pm$0.171 & 17$\pm$0.1 \\
\hline
\multirow{3}[6]{*}{GENE-CH} & $10^{-4}$ & 6.28$\pm$1.89 & 9.19$\pm$1.27 & 13.1$\pm$1.14 & 16.6$\pm$0.861 & 20.2$\pm$0.443 \\
\cline{2-7}      & $10^{-5}$ & 5.08$\pm$1.04 & 8.44$\pm$0.671 & 12.4$\pm$0.715 & 16.1$\pm$0.327 & 20$\pm$0.245 \\
\cline{2-7}      & $10^{-6}$ & 4.39$\pm$0.634 & 8.12$\pm$0.327 & 12.1$\pm$0.239 & 16$\pm$0.1 & 20$\pm$0.174 \\
\hline
\multirow{3}[6]{*}{GENE-AH} & $10^{-4}$ & 5.65$\pm$1.06 & 8.99$\pm$1.07 & 12.8$\pm$0.94 & 16.4$\pm$0.609 & 20$\pm$0.284 \\
\cline{2-7}      & $10^{-5}$ & 4.95$\pm$0.914 & 8.41$\pm$0.653 & 12.3$\pm$0.525 & 16.1$\pm$0.256 & 19.9$\pm$0.356 \\
\cline{2-7}      & $10^{-6}$ & 4.35$\pm$0.575 & 8.1$\pm$0.302 & 12$\pm$0.141 & 16$\pm$0 & 19.7$\pm$0.482 \\
\hline
\multirow{3}[6]{*}{HFC} & $10^{-4}$ & 4$\pm$0 & 5.93$\pm$0.293 & 9$\pm$0 & 8.14$\pm$0.817 & 3.01$\pm$0.1 \\
\cline{2-7}      & $10^{-5}$ & 4$\pm$0 & 5.82$\pm$0.386 & 9$\pm$0 & 7.24$\pm$0.495 & 2.99$\pm$0.1 \\
\cline{2-7}      & $10^{-6}$ & 4$\pm$0 & 5.93$\pm$0.541 & 9$\pm$0.239 & 8.14$\pm$0.284 & 3.01$\pm$0.171 \\
\hline
\multirow{3}[6]{*}{NWHFC} & $10^{-4}$ & 4$\pm$0 & 4.13$\pm$0.367 & 9.36$\pm$0.482 & 7.15$\pm$0.359 & 3.11$\pm$0.314 \\
\cline{2-7}      & $10^{-5}$ & 4$\pm$0 & 4$\pm$0 & 9.08$\pm$0.273 & 7.01$\pm$0.1 & 3.01$\pm$0.1 \\
\cline{2-7}      & $10^{-6}$ & 4$\pm$0 & 4$\pm$0 & 9$\pm$0 & 6.98$\pm$0.141 & 2.99$\pm$0.1 \\
\hline
\multirow{3}[6]{*}{ATGP-NPD} & $10^{-4}$ & 18.4$\pm$1.24 & 24.1$\pm$1.62 & 23.7$\pm$1.49 & 20.9$\pm$1.51 & 20.9$\pm$0.81 \\
\cline{2-7}      & $10^{-5}$ & 15.9$\pm$1.11 & 20.6$\pm$1.27 & 20.6$\pm$1.33 & 18.5$\pm$1.07 & 20.2$\pm$0.647 \\
\cline{2-7}      & $10^{-6}$ & 14.1$\pm$0.886 & 18.5$\pm$1.04 & 18$\pm$1.4 & 17.6$\pm$0.777 & 19.8$\pm$0.512 \\
\hline
\hline
\end{tabular}%
  \label{tab:MOS_Nvary}%
\end{table}%

\fi

\section{Real Data Experiment}

\begin{figure}
	\centering
	\includegraphics[width=.48\textwidth]{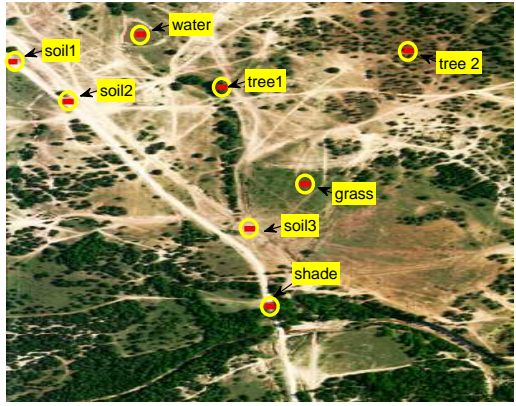}
	\caption{The RGB image of the TERRAIN HSI. The red dots are the visually picked pure pixels.}
	\label{fig:RGB}
\end{figure}

\begin{figure}
    \centering
    \includegraphics[width=.45\textwidth]{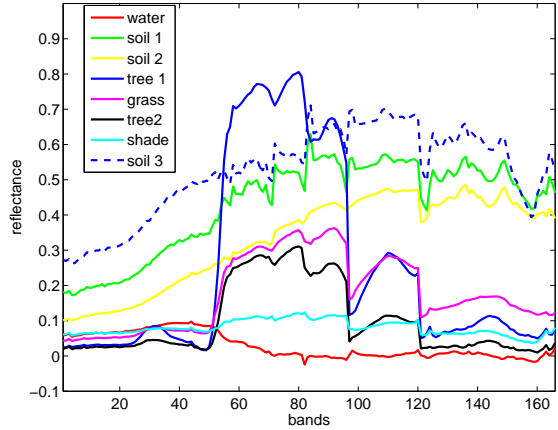}
    \caption{Spectra of visually picked pure pixels in the TERRAIN HSI experiment.} \label{fig:visual}
\end{figure}

In this section, we test $\ell_q$ SD-SOMP on the TERRAIN HSI, which is a $550\times 307$ image captured by the HYDICE sensor \cite{rickard1993hydice}.
The RGB image of this dataset is shown in Fig.~\ref{fig:RGB}.
Owing to its high spatial resolution, the number of endmembers can be visually checked to be around $5$--$8$;
in fact, the first five principal components of the dataset contain more than $99\%$ of the total energy.
Some visually identified pure pixels are plotted in Fig.~\ref{fig:visual}.
We remove highly contaminated bands, specifically, bands $1$--$4$, $76$, $87$, $103$--$111$, $136$--$153$ and $200$--$201$, which results in $166$ active bands.
Our experiment is conducted following the procedure suggested in \cite{GENE}:
1) reduce the dimension of the pixels to $N_{\max}=50$ by inexact ASF-DR, apply $\ell_\infty$ SD-SOMP to select a set of pixels $\hat{\Lambda}$, and let $\hat{N}=|\hat{\Lambda}|$;
2) apply \emph{exact} ASF-DR with dimension $\hat{N}$ for further suppression of noise, and then apply $\ell_\infty$ SD-SOMP to select $\hat{N}$ pure pixels;
3) transform the selected dimension-reduced pixels to the original space to obtain the estimated endmembers.

The estimated model orders by $\ell_\infty$ SD-SOMP, GENE-AH, GENE-CH and HYSIME are listed in Table~\ref{tab:real}.
In this experiment, GENE-CH fails to converge within its maximum number of iterations (i.e., 50), and GENE-AH and HYSIME yield $\hat{N}=42$ and $\hat{N}=24$, resp.
The estimated model order by the proposed $\ell_\infty$ SD-SOMP is $\hat{N}=7$, which is close to the number of the visually identified materials.

\begin{figure}
 \centering
  \includegraphics[width=.45\textwidth]{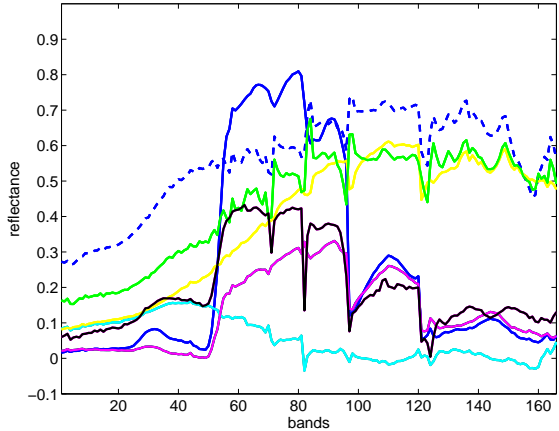}
  \caption{Estimated endmember spectra by $\ell_\infty$ SD-SOMP in the TERRAIN HSI experiment.} \label{fig:SOMP}
\end{figure}

\begin{figure}
    \centering
    \includegraphics[width=.45\textwidth]{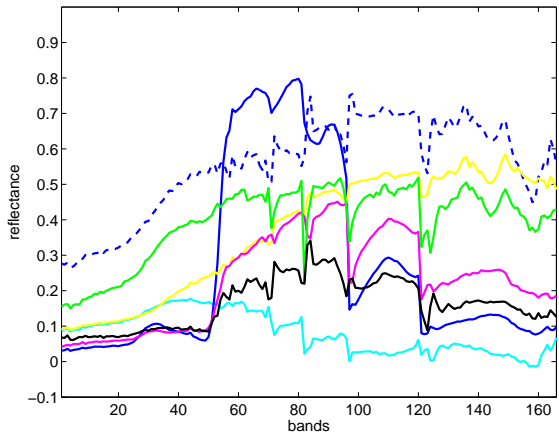}
    \caption{Estimated endmember spectra by VCA in the TERRAIN HSI experiment.} \label{fig:VCA}
\end{figure}

Fig.~\ref{fig:SOMP} shows the estimated endmember spectra by $\ell_\infty$ SD-SOMP.
To benchmark, we run another algorithm, namely, VCA \cite{VCA}.
VCA does not handle unknown model order, and we run it by prespecifying $\hat{N}=7$ (the same number as that identified by $\ell_\infty$ SD-SOMP).
The results are displayed in Fig.~\ref{fig:VCA}.
We see that both $\ell_\infty$ SD-SOMP and VCA miss `water'; this may be due to the fact that the spectrum of `water' is highly correlated with that of `shade' (cf. Fig.~\ref{fig:visual}).
To quantify the endmember estimation accuracy,
we use the mean-removed spectral angle (MRSA):
\[\bar{\phi}={\rm arccos}\left(\frac{({\bf a}_{\rm est}-{\bf m}({{\bf a}_{\rm est}}))^T({\bf a}_{\rm visual}-{\bf m}({{\bf a}_{\rm visual}}))}{\|{\bf a}_{\rm est}-{\bf m}({\bf a}_{\rm est})\|_2\|{\bf a}_{\rm visual}-{\bf m}({{\bf a}_{\rm visual}})\|_2} \right),\]
where ${\bf m}({\bf a})=(1/M)({\bf 1}^T{\bf a}){\bf 1}$, and
${\bf a}_{\rm visual}$ and ${\bf a}_{\rm est}$ denote the spectrum of the visually identified pure pixel and the corresponding best-matched spectrum of the pure pixel selected by
an
algorithm, respectively.
MRSA is the angle between the mean-removed estimated spectrum and the visually selected spectrum---a small MRSA means that the two spectra match with each other well.
The results, tabulated in Table~\ref{tab:MRSA}, indicate that $\ell_\infty$ SD-SOMP gives reasonably good MRSAs.

%

\begin{table}
	\centering
\footnotesize
	\caption{The estimated number of endmembers by various algorithms in the TERRAIN HSI experiment.}\label{tab:real}
			\begin{tabular}{c|c|c}
\hline
\hline
Algorithm & $P_{\rm FA}$   & estimated $N$ \\
\hline
GENE-AH  & $10^{-8}$ & 42 \\
\hline
GENE-CH & $10^{-8}$ & - \\
\hline
HYSIME & -     & 24 \\
\hline
$\ell_\infty$ SD-SOMP &  -     & 7 \\
\hline
\hline
\end{tabular}%
\end{table}


%

\begin{table}
	\centering
\caption{Mean-removed spectral angles $\bar{\phi}$ (degrees) of $\ell_\infty$ SD-SOMP and VCA in the TERRAIN HSI experiment.}\label{tab:MRSA}
\begin{tabular}{c|c|c|c|c|c|c|c}
\hline
\hline
Algorithm & soil 1 & soil 2 & soil 3 & shade & tree 1 & tree 2 & grass \\
\hline
\hline
VCA  & 6.95  & 4.49  & 1.74  & 33.10  & 3.53  & 31.24  & 7.72  \\
\hline
$\ell_\infty$ SOMP  & 4.00  & 4.92  & 1.64  & 47.20  & 0.67  & 21.43  & 16.73  \\
\hline
\hline
\end{tabular}%
\end{table}

\section{Conclusion}
In this paper, we have considered a greedy approach for handling the self-dictionary sparse regression formulation for hyperspectral unmixing.
The relationship between greedy self-dictionary sparse regression and pure pixel search was revealed, and exact recovery conditions of the proposed greedy algorithm in the noisy case were analyzed.
Numerical results and an experiment based on the TERRAIN HSI dataset showed that under the pure pixel assumption, the proposed greedy algorithm yields good
endmember and model-order identification performance.


\appendix

\ifplainver
    \section*{Appendix}
    \renewcommand{\thesubsection}{\Alph{subsection}}
\else
    \section{Appendix}
\fi

\subsection{Proof of Theorem~\ref{thm:SD-SOMP} and Observation~\ref{obs:SD-SOMP}}
\label{sec:proof_SD-SOMP}

Our proof takes insight from that of \cite[Property 3]{VMAX}.
In particular, we use induction.
Recall the greedy selection step at the $k$th iteration of $\ell_q$ SD-SOMP in Algorithm~\ref{thm:SD-SOMP}:
\begin{equation} \label{eq:new_greedy}
\hat{n}_k = \arg \max_{n=1,\ldots,L} \| {\bf R}_{k-1}^T {\bf x}[n] \|_q,
\end{equation}
where $1 \leq k \leq N$ and ${\bf R}_{k-1} = {\bf P}^\perp_{{\bf X}_{\Lambda_{k-1}}} {\bf X}$.
Suppose that $\hat{n}_i$ is a pure pixel index of endmember $i$ for $i \in \{1,\ldots,k-1\}$.
By observing ${\bf X}_{\Lambda_{k-1}} = [~ {\bf a}_1, \ldots, {\bf a}_{k-1} ~] = {\bf A}_{1:k-1}$ in the noiseless case, we
have
\begin{subequations} \label{eq:t1}
\begin{align}
\| {\bf R}_{k-1}^T {\bf x}[n] \|_q
& = \left\| \sum_{i=k}^N {\bf X}^T {\bf P}^\perp_{{\bf A}_{1:k-1}} {\bf a}_i s_i[n]  \right\|_q  \label{eq:t1a} \\
&  \leq
\sum_{i=k}^N s_i[n] \left\|  {\bf X}^T {\bf P}^\perp_{{\bf A}_{1:k-1}} {\bf a}_i   \right\|_q  \label{eq:t1b}  \\
& \leq \max_{i=k,\ldots,N} \left\|  {\bf X}^T {\bf P}^\perp_{{\bf A}_{1:k-1}} {\bf a}_i   \right\|_q,  \label{eq:t1c}
\end{align}
\end{subequations}
where $s_i[n]$ denotes the $i$th element of ${\bf s}[n]$,
\eqref{eq:t1b} is obtained by applying the triangle inequality,
and \eqref{eq:t1c} by the properties ${\bf s}[n] \geq {\bf 0}$, ${\bf 1}^T {\bf s}[n] = 1$.
Let us assume $\|  {\bf X}^T {\bf P}^\perp_{{\bf A}_{1:k-1}} {\bf a}_k   \|_q = \max_{i=k,\ldots,N} \|  {\bf X}^T {\bf P}^\perp_{{\bf A}_{1:k-1}} {\bf a}_i  \|_q$.
It is seen that equality in \eqref{eq:t1} holds when ${\bf s}[n] = {\bf e}_k$.
Hence, the maximum in
\eqref{eq:new_greedy}
is attained when $n$ is a pure pixel index of endmember $k$.

We also need to show that the maximum in
\eqref{eq:new_greedy}
is attained only when $n$ is a pure pixel index.
Consider the case of $1 < q < \infty$.
By the Minkowski inequality, equality in \eqref{eq:t1b} holds only if
\begin{equation} \label{eq:t3}
{\bf X}^T {\bf P}^\perp_{{\bf A}_{1:k-1}} {\bf a}_i = {\bf X}^T {\bf P}^\perp_{{\bf A}_{1:k-1}} {\bf a}_j
\end{equation}
for all $i,j \in \{ k,\ldots,N \}$, $i \neq j$, with $s[i] \neq 0$, $s[j] \neq 0$.
However, \eqref{eq:t3} does not hold:
The matrix ${\bf X}$ can be shown to have full column rank when $\{ {\bf a}_1, \ldots, {\bf a}_N \}$ is linearly independent and the pure pixel assumption holds.
As a result, \eqref{eq:t3} is equivalent to ${\bf P}^\perp_{{\bf A}_{1:k-1}} {\bf a}_i = {\bf P}^\perp_{{\bf A}_{1:k-1}} {\bf a}_j$, which does not hold when  $\{ {\bf a}_1, \ldots, {\bf a}_N \}$ is linearly independent.
We therefore conclude that the greedy selection step
in \eqref{eq:new_greedy}
must choose
a pure pixel index of a previously unidentified endmember.

The proof above does not cover the case of $q = \infty$.
For $q = \infty$, the greedy selection step can be expressed as
\ifconfver
    \begin{subequations} \label{eq:t4}
    \begin{align}
    & \max_{n=1,\ldots,L} \left\| {\bf R}_{k-1}^T {\bf x}[n] \right\|_\infty  \nonumber \\
    &   =
    \max_{n=1,\ldots,L} \max_{m=1,\ldots,L} \left| {\bf x}^T[m] {\bf P}_{{\bf X}_{\Lambda_{k-1}}}^\perp {\bf x}[n] \right| \label{eq:t4a} \\
    & = \max_{n=1,\ldots,L} \max_{m=1,\ldots,L} \left| ({\bf P}_{{\bf X}_{\Lambda_{k-1}}}^\perp {\bf x}[m] )^T ( {\bf P}_{{\bf X}_{\Lambda_{k-1}}}^\perp {\bf x}[n] ) \right|  \label{eq:t4b} \\
    & \leq \max_{n=1, \ldots, L} \| {\bf P}_{{\bf X}_{\Lambda_{k-1}}}^\perp {\bf x}[n] \|_2^2 \label{eq:t4c}
    \end{align}
    \end{subequations}
\else
    \begin{subequations} \label{eq:t4}
    \begin{align}
    \max_{n=1,\ldots,L} \left\| {\bf R}_{k-1}^T {\bf x}[n] \right\|_\infty  &   =
    \max_{n=1,\ldots,L} \max_{m=1,\ldots,L} \left| {\bf x}^T[m] {\bf P}_{{\bf X}_{\Lambda_{k-1}}}^\perp {\bf x}[n] \right| \label{eq:t4a} \\
    & = \max_{n=1,\ldots,L} \max_{m=1,\ldots,L} \left| ({\bf P}_{{\bf X}_{\Lambda_{k-1}}}^\perp {\bf x}[m] )^T ( {\bf P}_{{\bf X}_{\Lambda_{k-1}}}^\perp {\bf x}[n] ) \right|  \label{eq:t4b} \\
    & \leq \max_{n=1, \ldots, L} \| {\bf P}_{{\bf X}_{\Lambda_{k-1}}}^\perp {\bf x}[n] \|_2^2 \label{eq:t4c}
    \end{align}
    \end{subequations}
\fi
where \eqref{eq:t4c} is due to the Cauchy-Schwartz inequality.
Moreover, equality in \eqref{eq:t4c} is attained when we choose $m= n$ in \eqref{eq:t4a}.
Hence, the greedy selection step can be simplified to $\hat{n}_k = \arg \max_{n=1,\ldots,L} \| {\bf P}_{{\bf X}_{\Lambda_{k-1}}}^\perp {\bf x}[n] \|_2^2$.
The resulting algorithm takes the same form as SPA~\cite{Gillis2012},
which has been proven to yield exact pure pixel identifiability \cite[Property~3]{VMAX} (see also \cite[Theorem~1]{Gillis2012}  and \cite{Ma2014}).

\subsection{Proof of Lemma~\ref{lem:1}}\label{app:lem1}
Let us assume w.l.o.g. that each $k \in \{ 1, \ldots, N \}$ is
a
pure pixel index of endmember $k$, so that we can conveniently write
${\bf x}[k] = {\bf a}_k + {\bf v}[k], k=1,\ldots,N.$
Consider the following ${\bf C}$
\begin{equation} \label{eq:lem:1:C}
{\bf C}= \begin{bmatrix} {\bf s}[1] & {\bf s}[2] & \ldots & {\bf s}[L] \\
                            {\bf 0} & {\bf 0} & \ldots & {\bf 0}
\end{bmatrix},
\end{equation}
where ${\bf s}[n] \in \mathbb{R}^N$, $n=1,\ldots,L$, are the true abundance vectors.
Clearly, such a ${\bf C}$ satisfies \eqref{eq:SD-MMV-noisy_c}.
It also satisfies \eqref{eq:SD-MMV-noisy_b} for $\delta \geq 2 \epsilon$, since
\begin{align*}
\| {\bf x}[n] - {\bf X}{\bf c}_n \|_2 & = \left\| {\bf A}{\bf s}[n] + {\bf v}[n] - \sum_{i=1}^N s_i[n] ( {\bf a}_i + {\bf v}[i] ) \right\|_2 \\
& = \left\| {\bf v}[n] - \sum_{i=1}^N s_i[n] {\bf v}[i]  \right\|_2 \\
& \leq \| {\bf v}[n] \|_2 + \sum_{i=1}^N s_i[n] \| {\bf v}[i] \|_2 \\
& \leq 2 \epsilon \leq \delta
\end{align*}
for any $n \in \{ 1,\ldots, L \}$.
Note that to obtain the above inequality, we have used the basic abundance assumptions ${\bf s}[n] \geq {\bf 0},$ ${\bf 1}^T {\bf s}[n] =1$.
Thus, the ${\bf C}$ in \eqref{eq:lem:1:C} is a feasible point of Problem~\eqref{eq:SD-MMV-noisy}.
It also follows from \eqref{eq:lem:1:C} that $\| {\bf C} \|_{\rm row-0} = N$.

\subsection{Proof of Lemma~\ref{lem:2}}\label{app:lemma2}
We prove Lemma~\ref{lem:2} by contradiction.
Denote $r = 2(\delta + 2 \epsilon)/\sigma_{\rm min}({\bf A})$ and $\Lambda = {\rm rowsupp}({\bf C})$ for convenience.
Suppose that \eqref{eq:lem:2:bound} does not hold; this means that
\begin{equation} \label{eq:lem:2:proof_cond1}
\| {\bf e}_k  - {\bf s}[\hat{n}] \|_1 > r, ~ \forall \hat{n} \in \Lambda,
\end{equation}
where $k \in \{ 1, \ldots, N \}$.
Our task is to show that \eqref{eq:lem:2:proof_cond1} cannot be satisfied.
To proceed, first note the following identity for any ${\bf s} \geq {\bf 0}$, ${\bf 1}^T {\bf s} = 1$:
\begin{align}
\| {\bf e}_k - {\bf s} \|_1 & = ( 1 - s_k ) + \sum_{i \neq k} s_i \nonumber \\
& = 2 ( 1- s_k ). \label{eq:lem:2:proof_useful}
\end{align}
Also, assume w.l.o.g. that $k \in \{ 1,\ldots, N \}$ is a pure pixel index of endmember $k$.
Then,
we have
\begin{align}
\| {\bf x}[k] - {\bf X}{\bf c}_k \|_2
   & = \left\| {\bf a}_k + {\bf v}[k] - \sum_{i \in \Lambda} c_{k,i} ( {\bf A}{\bf s}[i] + {\bf v}[i] ) \right\|_2 \nonumber \\
   & \geq \alpha - \beta, \label{eq:lem:2:proof_cond2}
\end{align}
where
\[ \alpha = \left\| {\bf A} \left( {\bf e}_k - \sum_{i \in \Lambda} c_{k,i} {\bf s}[i] \right) \right\|_2,
\quad
\beta= \left\| {\bf v}[k] - \sum_{i \in \Lambda} c_{k,i} {\bf v}[i] \right\|_2.
\]
Let us further derive a bound on \eqref{eq:lem:2:proof_cond2}.
Using the feasibility of ${\bf C}$ w.r.t. Problem~\eqref{eq:SD-MMV-noisy}, specifically, ${\bf c}_k \geq {\bf 0}$, ${\bf 1}^T {\bf c}_k = 1$, one can easily verify that
\begin{equation} \label{eq:lem:2:proof_beta}
\beta \leq \| {\bf v}[k] \|_2 + \sum_{i \in \Lambda} c_{k,i} \| {\bf v}[i] \|_2 \leq 2 \epsilon.
\end{equation}
Moreover, the following inequality can be shown
\begin{subequations} \label{eq:lem:2:proof_cond3}
\begin{align}
\alpha & \geq
    \sigma_{\rm min}({\bf A}) \left\| {\bf e}_k - \sum_{i \in \Lambda} c_{k,i} {\bf s}[i]  \right\|_2    \\
    &  \geq     \sigma_{\rm min}({\bf A}) \left| 1 - \sum_{i \in \Lambda} c_{k,i} s_k[i] \right|
                   \label{eq:lem:2:proof_cond3b} \\
    &  \geq     \sigma_{\rm min}({\bf A}) \left( 1 - \max_{i \in \Lambda} s_k[i] \right)
               \label{eq:lem:2:proof_cond3c} \\
    &  >     \sigma_{\rm min}({\bf A})  \cdot \frac{r}{2} = \delta + 2 \epsilon,
               \label{eq:lem:2:proof_cond3d}
\end{align}
\end{subequations}
where \eqref{eq:lem:2:proof_cond3b} is by $\| {\bf x} \|_2 \geq |x_k|$ for any $k$;
\eqref{eq:lem:2:proof_cond3c} is by ${\bf c}_k \geq {\bf 0}$, ${\bf 1}^T {\bf c}_k = 1$, and $1 \geq s_k[i] \geq 0$ for all $i$;
\eqref{eq:lem:2:proof_cond3d} is by \eqref{eq:lem:2:proof_cond1}, \eqref{eq:lem:2:proof_useful}, and the definition of $r$.
Substituting \eqref{eq:lem:2:proof_beta} and \eqref{eq:lem:2:proof_cond3} into \eqref{eq:lem:2:proof_cond2} leads to
\[ \| {\bf x}[k] - {\bf X}{\bf c}_k \|_2 > \delta, \]
which violates the feasibility condition
\eqref{eq:SD-MMV-noisy_b}.
Hence, we have proven that \eqref{eq:lem:2:proof_cond1} contradicts with the feasibility of ${\bf C}$.
As a result,
the desired result in \eqref{eq:lem:2:bound} must hold.

We should also consider the case of $r < 1$.
Using \eqref{eq:lem:2:proof_useful}, we can re-express
\eqref{eq:lem:2:bound} as
\begin{equation}   \label{eq:lem:2:proof_cond5}
s_k[\hat{n}_k] \geq 1 - \frac{r}{2} > \frac{1}{2}, \quad k=1,\ldots,N.
\end{equation}
Suppose that some of the $\hat{n}_1, \ldots, \hat{n}_N $ are repeated; i.e., $\hat{n}_k = \hat{n}_j$ for some $k \neq j$.
Then, one can verify from \eqref{eq:lem:2:proof_cond5} that ${\bf 1}^T {\bf s}[\hat{n}_k] > 1$, a contradiction to the abundance sum-to-one assumption ${\bf 1}^T {\bf s}[n] = 1$.
Hence, we conclude that $r < 1$ implies distinct $\hat{n}_1, \ldots, \hat{n}_N $.

\subsection{ Proof of Theorem~\ref{thm:SD-MMV-noisy}} \label{app:SD-MMV-noisy}

Step 1: \ Suppose that
\begin{equation}  \label{eq:thm:SD-MMV-noisy:cond0}
\delta \geq 2 \epsilon, \quad \frac{2(\delta + 2 \epsilon)}{\sigma_{\rm min}({\bf A})} < 1
\end{equation}
are satisfied simultaneously.
Then, by Lemmas~\ref{lem:1}-\ref{lem:2}, an optimal solution ${\bf C}_{\rm opt}$ to Problem~\eqref{eq:SD-MMV-noisy} must satisfy i) $\| {\bf C}_{\rm opt} \|_{\rm row-0} = N$, and ii)
\begin{equation} \label{eq:thm:SD-MMV-noisy:cond00}
\| {\bf e}_k - {\bf s}[\hat{n}_k] \|_1 \leq \frac{2(\delta + 2 \epsilon)}{\sigma_{\rm min}({\bf A})},
~ \text{for $k=1,\ldots,N$,}
\end{equation}
where we denote ${\rm rowsupp}({\bf C}_{\rm opt}) = \{ \hat{n}_1, \ldots, \hat{n}_N \}$.
Equations~\eqref{eq:thm:SD-MMV-noisy:cond0} can be shown to be satisfied when
\begin{align}
\epsilon  < \frac{\sigma_{\rm min}({\bf A})}{8},  & \quad
2 \epsilon \leq \delta < \frac{\sigma_{\rm min}({\bf A})}{2} - 2 \epsilon. \label{eq:thm:SD-MMV-noisy:cond02}
\end{align}
The latter equation leads to the condition on $\delta$ in Theorem~\ref{thm:SD-MMV-noisy}.

Step 2: \ Let us further consider a case where
\begin{equation} \label{eq:thm:SD-MMV-noisy:cond1}
\| {\bf e}_k - {\bf s}[\hat{n}_k] \|_1 < d({\bf S}),
\end{equation}
for all $k \in \{1,\ldots,N \}$. By the definition of $d({\bf S})$, one can easily verify that \eqref{eq:thm:SD-MMV-noisy:cond1} is identical to
${\bf s}[\hat{n}_k] = {\bf e}_k$; i.e., $\{ \hat{n}_1, \ldots, \hat{n}_N \}$ is a complete pure pixel index set.
Equations~\eqref{eq:thm:SD-MMV-noisy:cond1} are achieved when
\begin{equation} \label{eq:thm:SD-MMV-noisy:cond15}
\frac{2(\delta + 2 \epsilon)}{\sigma_{\rm min}({\bf A})} < d({\bf S}).
\end{equation}
By combining \eqref{eq:thm:SD-MMV-noisy:cond02} and \eqref{eq:thm:SD-MMV-noisy:cond15}, we obtain the sufficient exact recovery condition \eqref{eq:exact_reov} in Theorem~\ref{thm:SD-MMV-noisy}.

\subsection{ Proof of Corollary~\ref{cor:SD-MMV-noisy}} \label{app:cor-SD-MMV-noisy}
The proof of Corollary~\ref{cor:SD-MMV-noisy} is the same as that of Theorem~\ref{thm:SD-MMV-noisy} shown above,
except that we do not consider Step 2 there.
Additionally,
the error bound in Corollary~\ref{cor:SD-MMV-noisy} is obtained by
\begin{subequations}
\begin{align}
\| {\bf a}_k - {\bf x}[\hat{n}_k] \|_2 & \leq \| {\bf A}({\bf e}_k - {\bf s}[\hat{n}_k]) \|_2 + \epsilon \\
& \leq \left( \max_{i=1,\ldots,N} \| {\bf a}_i \|_2 \right) \| {\bf e}_k - {\bf s}[\hat{n}_k] \|_1 + \epsilon
\label{eq:cor:SD-MMV-noisy:cond1b} \\
& \leq \frac{\displaystyle 2(\delta+2\epsilon) \left( \max_{i=1,\ldots,N} \| {\bf a}_i \|_2 \right)  }{\sigma_{\rm min}({\bf A})} + \epsilon, \label{eq:cor:SD-MMV-noisy:cond1c}
\end{align}
\end{subequations}
where \eqref{eq:cor:SD-MMV-noisy:cond1b} is due to the triangle inequality,
and \eqref{eq:cor:SD-MMV-noisy:cond1c} is by \eqref{eq:thm:SD-MMV-noisy:cond00}.

\subsection{Proof of Theorem~\ref{thm:SOMP1}} \label{app:thm-SOMP1}
The proof is divided into four steps.

Step 1: \ First, we show that $\{ \hat{n}_1,\ldots,\hat{n}_N \}$ is a complete pure pixel index set under certain conditions on $\epsilon$.
Suppose that condition \eqref{eq:SPA_eps_bnd} in Fact~\ref{fact:SPA} is satisfied, and
assume w.l.o.g. that the permutation $\bm \pi$ in \eqref{eq:SPA_e_bnd} is $\bm \pi = (1,2,\ldots,N)$.
From \eqref{eq:SPA_e_bnd}, we get
\begin{align}
\epsilon \cdot \eta({\bf A}) & \geq \| {\bf a}_k - {\bf x}[\hat{n}_k] \|_2 \nonumber \\
& \geq \frac{ \sigma_{\rm min}({\bf A}) }{ 2 } \| {\bf e}_k - {\bf s}[\hat{n}_k] \|_1 - \epsilon
\label{eq:thm:SOMP1:t1}
\end{align}
where \eqref{eq:thm:SOMP1:t1} is derived by the same way as in the proof of Lemma~\ref{lem:2},
particularly, \eqref{eq:lem:2:proof_cond2}-\eqref{eq:lem:2:proof_cond3}.
Let us reorganize \eqref{eq:thm:SOMP1:t1} as
\begin{equation}
\| {\bf e}_k - {\bf s}[\hat{n}_k] \|_1 \leq \frac{ 2 ( 1 + \eta({\bf A}) ) \epsilon}{ \sigma_{\rm min}({\bf A}) }
\triangleq r.
\label{eq:thm:SOMP1:t2}
\end{equation}
We note that $ {\bf s}[\hat{n}_k]  = {\bf e}_k $ if $r < d({\bf S})$ (see the definition of $d({\bf S})$ in \eqref{eq:dS}).
The latter condition is shown to be satisfied when
\begin{equation}
\epsilon  < \frac{ \sigma_{\rm min}({\bf A}) \cdot  d({\bf S}) }{ 2 ( 1 + \eta({\bf A}) ) }
\leq \frac{ \sigma_{\rm min}({\bf A}) \cdot  d({\bf S}) }{ 4 \eta({\bf A})  },
\label{eq:thm:SOMP1:cond1}
\end{equation}
where we have used $\eta({\bf A}) \geq 1$ to obtain the second inequality.

Step 2: \ Second, we examine conditions under which Stopping Rule 1 does not hold for any iteration number $k \leq N-1$.
To proceed, assume w.l.o.g. that each index $j \in \{ 1,\ldots, N \}$ is a pure pixel index of endmember $j$, and that $\hat{n}_j = j$ for all $j \in \{ 1,\ldots,N \}$.
With this setting, we can write ${\bf x}[i]= {\bf a}_{i} + {\bf v}[i]$ for all $i=1,\ldots,N$ and ${\bf X}_{\Lambda_k} = {\bf A}_{1:k} + {\bf V}_{1:k}$.
Consequently, the objective function in \eqref{eq:e_n} for $n=k+1$ is shown to yield a lower bound
\begin{align}
\| {\bf x}[k+1] - {\bf X}_{\Lambda_k} \bar{\bf c} \|_2
   & \geq
     \sigma_{\rm min}({\bf A})
      \left\| \begin{bmatrix} {\bf 0} \\ 1 \\ {\bf 0}  \end{bmatrix} -  \begin{bmatrix} \bar{\bf c} \\ 0 \\ {\bf 0}  \end{bmatrix} \right\|_2 - 2 \epsilon \nonumber \\
    & \geq \sigma_{\rm min}({\bf A})  - 2 \epsilon,
    \label{eq:thm:SOMP1:t3}
\end{align}
for any $\bar{\bf c} \geq {\bf 0}$, ${\bf 1}^T \bar{\bf c} = 1$,
where the proof of the above inequality is analogous to \eqref{eq:lem:2:proof_cond2}-\eqref{eq:lem:2:proof_cond3} in the proof of Lemma~\ref{lem:2}.
We see from \eqref{eq:thm:SOMP1:t3} and \eqref{eq:e_n} that Stopping Rule 1 is not satisfied if
\begin{equation}  \label{eq:thm:SOMP1:cond2}
\sigma_{\rm min}({\bf A})  - 2 \epsilon > \delta.
\end{equation}

Step 3: \
Third, we identify a condition under which Stopping Rule 1 is satisfied at iteration $k=N$.
Under the same setting as the previous step, we have ${\bf X}_{\Lambda_N} = {\bf A} + {\bf V}_{1:N}$.
From \eqref{eq:e_n}, it can be easily verified that
\begin{subequations} \label{eq:thm:SOMP1:t4}
\begin{align}
\min_{ \bar{\bf c} \geq {\bf 0}, {\bf 1}^T \bar{\bf c}= 1 }  \| {\bf x}[n]- {\bf X}_{\Lambda_N} \bar{\bf c} \|_2
   & \leq \| {\bf x}[n]- {\bf X}_{\Lambda_N} {\bf s}[n] \|_2 \label{eq:thm:SOMP1:t4a} \\
   & \leq 2 \epsilon, \label{eq:thm:SOMP1:t4b}
\end{align}
\end{subequations}
for any $n \in \{ 1,\ldots,L \}$,
where
\eqref{eq:thm:SOMP1:t4b} is obtained by the same proof as in Lemma~\ref{lem:1}.
Equation~\eqref{eq:thm:SOMP1:t4} suggests that Stopping Rule 1 is satisfied at $k=N$ if $2 \epsilon \leq \delta$ holds.

Step 4: \
Last, we combine the conditions obtained in Steps 1-3, namely, \eqref{eq:SPA_eps_bnd}, \eqref{eq:thm:SOMP1:cond1}, \eqref{eq:thm:SOMP1:cond2}, and $\delta \geq 2 \epsilon$, for achieving  a complete $\hat{\Lambda}$.
The aforementioned equations are shown to hold simultaneously if \eqref{eq:SOMP1_e_bnd} and $\delta \in [ 2\epsilon, \sigma_{\rm min}({\bf A})- 2\epsilon)$ are true.
Theorem~\ref{thm:SOMP1} is therefore proven.

\subsection{Proof of Theorem~\ref{thm:SOMP2}} \label{app:thm-SOMP2}
The proof is almost the same as that of Theorem~\ref{thm:SOMP1}.
The only difference is that in Step 2 of Theorem~\ref{thm:SOMP1}, we should
replace $\| {\bf x}[k+1] - {\bf X}_{\Lambda_k} \bar{\bf c} \|_2$ at the LHS of \eqref{eq:thm:SOMP1:t3} by
$\| {\bf x}[\hat{n}_{k+1}] - {\bf X}_{\Lambda_k} \bar{\bf c} \|_2$.
However, since the previous step has shown that $\hat{n}_{k+1} = k+1$, the proof turns out to have no difference.

\subsection{Proof of Corollary~\ref{cor:SOMP}} \label{app:cor-SOMP}
The proof is similar to that of Theorem~\ref{thm:SOMP1}. We concisely describe the proof by highlighting the key steps.
First, suppose that condition \eqref{eq:SPA_eps_bnd} in Fact~\ref{fact:SPA} holds.
Following Step 1 of the proof of Theorem~\ref{thm:SOMP1}, we have \eqref{eq:thm:SOMP1:t2}.
Applying \eqref{eq:lem:2:proof_useful} to \eqref{eq:thm:SOMP1:t2}, we further obtain
\begin{align}
s_k[n_k] & \geq 1 - \frac{r}{2}, \label{eq:cor:SOMP:t1} \\
s_i[n_k] & \leq \frac{r}{2}, \quad \text{for all $i \neq k$.} \label{eq:cor:SOMP:t2}
\end{align}
Second, consider Stopping Rule 1 for iteration $k \leq N-1$.
By the same setting as Step 2 of the proof of Theorem~\ref{thm:SOMP1},
we show that
\begin{subequations}  \label{eq:cor:SOMP:t3}
\begin{align}
\| {\bf x}[k+1] - {\bf X}_{\Lambda_k} \bar{\bf c} \|_2
	& \geq  \sigma_{\rm min}({\bf A}) \left\| {\bf e}_{k+1} - \sum_{i=1}^k \bar{c}_i {\bf s}[\hat{n}_i] \right\|_2 - 2\epsilon \label{eq:cor:SOMP:t3a}  \\
	& \geq \sigma_{\rm min}({\bf A}) \left| 1 -  \sum_{i=1}^k \bar{c}_i
s_{k+1}[\hat{n}_i]
\right| - 2\epsilon \label{eq:cor:SOMP:t3b} \\
	& \geq \sigma_{\rm min}({\bf A}) \left( 1 - \frac{r}{2} \right) - 2\epsilon, \label{eq:cor:SOMP:t3c} \\
	& = \sigma_{\rm min}({\bf A}) - (3 + \eta({\bf A}) ) \epsilon,
\end{align}
\end{subequations}
for any $\bar{\bf c} \geq {\bf 0}$, ${\bf 1}^T \bar{\bf c} = 1$,
where \eqref{eq:cor:SOMP:t3a} is obtained by the same way as in \eqref{eq:lem:2:proof_cond2}-\eqref{eq:lem:2:proof_cond3};
\eqref{eq:cor:SOMP:t3b} is by $\| {\bf x} \|_2 \geq |x_i|$ for any $i$;
\eqref{eq:cor:SOMP:t3c} is by \eqref{eq:cor:SOMP:t2}.
Hence, if
\begin{equation}  \label{eq:cor:SOMP:u1}
\sigma_{\rm min}({\bf A}) - (3 + \eta({\bf A}) ) \epsilon > \delta,
\end{equation}
then $\| {\bf x}[k+1] - {\bf X}_{\Lambda_k} \bar{\bf c} \|_2 \leq \delta$ does not hold.
Consequently, Stopping Rule 1 is not satisfied.
Third, consider Stopping Rule 1 at iteration $k=N$.
By the triangle inequality and Fact~\ref{fact:SPA},
it can be shown that
\begin{align}
\min_{ \bar{\bf c} \geq {\bf 0}, {\bf 1}^T \bar{\bf c}= 1 }  \| {\bf x}[n]- {\bf X}_{\Lambda_N} \bar{\bf c} \|_2
    &  \leq \| {\bf x}[n] - {\bf X}_{\Lambda_N} {\bf s}[n] \|_2 \nonumber \\
	& = \left\| \sum_{i=1}^N s_i[n] ( {\bf a}_i - {\bf x}[\hat{n}_i] ) + {\bf v}[n] \right\|_2 \nonumber \\
	& \leq \epsilon \cdot \eta({\bf A}) + \epsilon, \label{eq:cor:SOMP:t4} \nonumber
\end{align}
for any $n \in \{ 1, \ldots, L \}$.
Hence, Stopping Rule 1 is satisfied at $k=N$ if
\begin{equation}  \label{eq:cor:SOMP:u2}
( 1 + \eta({\bf A}) ) \epsilon \leq \delta.
\end{equation}
Last, we find conditions under which \eqref{eq:SPA_eps_bnd}, \eqref{eq:cor:SOMP:u1} and \eqref{eq:cor:SOMP:u2} are satisfied simultaneously.
It is shown that the aforementioned equations hold if \eqref{eq:cor:SOMP:eps_bnd} is satisfied and $\delta$ lies in the range $[(1+\eta({\bf A})) \epsilon, \sigma_{\rm min}({\bf A})- (3 + \eta({\bf A})) \epsilon )$.

The above proof assumes Stopping Rule 1.
The proof for Stopping Rule 2 is the same, except for \eqref{eq:cor:SOMP:t3} where one should  consider $\| {\bf x}[\hat{n}_{k+1}] - {\bf X}_{\Lambda_k} \bar{\bf c} \|_2$.
By the same proof method as before, we prove that
\ifconfver
\begin{align*}
    & \| {\bf x}[\hat{n}_{k+1}]  - {\bf X}_{\Lambda_k} \bar{\bf c} \|_2 \\
    & \quad \geq \sigma_{\rm min}({\bf A}) \left| s_{k+1}[\hat{n}_{k+1}] -  \sum_{i=1}^k \bar{c}_i 
s_{k+1}[\hat{n}_i]
\right| - 2\epsilon
	\\
	& \quad \geq \sigma_{\rm min}({\bf A}) \left( 1 - r \right) - 2\epsilon,
	 \\
	& \quad  = \sigma_{\rm min}({\bf A}) - (4 + 2 \eta({\bf A}) ) \epsilon.
\end{align*}
\else
\begin{align*}
    \| {\bf x}[\hat{n}_{k+1}]  - {\bf X}_{\Lambda_k} \bar{\bf c} \|_2
	& \geq \sigma_{\rm min}({\bf A}) \left| s_{k+1}[\hat{n}_{k+1}] -  \sum_{i=1}^k \bar{c}_i 
s_{k+1}[\hat{n}_i]
\right| - 2\epsilon
	\\
	& \geq \sigma_{\rm min}({\bf A}) \left( 1 - r \right) - 2\epsilon,
	 \\
	& = \sigma_{\rm min}({\bf A}) - (4 + 2 \eta({\bf A}) ) \epsilon.
\end{align*}
\fi
By replacing the LHS of \eqref{eq:cor:SOMP:u1} with the above equation, the noise bound in \eqref{eq:cor:SOMP:eps_bnd2} and the corresponding range of $\delta$ are shown.

\bibliography{refs_mmv}

\end{document}